\newif\ifArXiV
\ArXiVtrue
\ifArXiV
\documentclass[10pt,letterpaper]{article}
\else
\documentclass[10pt,twocolumn,letterpaper]{article}
\fi
\usepackage[pagenumbers]{cvpr} 

\usepackage{graphicx}
\usepackage{amsmath}
\usepackage{amssymb}
\usepackage{booktabs}
\usepackage{MnSymbol}
\usepackage{tikz}
\usetikzlibrary{cd}
\usepackage[percent]{overpic}
\usepackage[utf8]{inputenc}
\usepackage{mathtools,siunitx}
\usepackage[shortlabels]{enumitem}
\usepackage{scalerel}

\usepackage[pagebackref,breaklinks,colorlinks]{hyperref}

\usepackage{amsmath,amssymb,amsfonts,hyperref,multirow,amsthm}
\usepackage{listings,etoolbox}

\newcommand{\RR}{\mathbb{R}}

\newcommand{\CC}{\mathbb{C}}
\newcommand{\arxiv}[1]{{#1}}

\newcommand{\hide}[1]{}
\newcommand{\tim}[1]{\textcolor{blue}{TD: #1}}
\newcommand{\anton}[1]{\textcolor{magenta}{AL: #1}}
\newcommand{\TP}[1]{\textcolor{red}{TP: #1}}
\newcommand{\PH}[1]{\textcolor{green}{PH: #1}}
\newcommand{\AScr}{A^{\scaleto{Scr}{4.5pt}}}
\newcommand{\Afp}{A^{\scaleto{5pt}{5.5pt}}}
\newcommand{\DScr}{D^{\scaleto{Scr}{4.5pt}}}
\newcommand{\Dfp}{D^{\scaleto{5pt}{5.5pt}}}
\newcommand{\PScr}{P^{\scaleto{Scr}{4.5pt}}}
\newcommand{\Pfp}{P^{\scaleto{5pt}{5.5pt}}}
\newcommand{\VScr}{V^{\scaleto{Scr}{4.5pt}}}
\newcommand{\Vfp}{V^{\scaleto{5pt}{5.5pt}}}
\DeclareMathOperator{\SE}{\rm SE}
\newtheorem{example}{Example}
\newtheorem{remark}{Remark}
\newcommand{\Mfpp}{M^{\scaleto{5pt}{5.5pt}}}
\newcommand{\MScr}{M^{\scaleto{Scr}{5.5pt}}}
\newcommand{\Psifpp}{\Psi^{\scaleto{5pt}{5.5pt}}}
\newcommand{\calS}{\mathcal{S}}
\newcommand{\PsiScr}{\Psi^{\scaleto{Scr}{5.5pt}}}
\DeclareMathOperator{\im}{\rm im}

\usepackage{xcolor}
\definecolor{codedarkpurple}{RGB}{179, 0, 89}
\definecolor{codeblue}{RGB}{51, 133, 255}
\definecolor{codelightpurple}{RGB}{255, 102, 217}
\lstdefinestyle{mystyle}{
basicstyle=\scriptsize\ttfamily,
xleftmargin=2em,
xrightmargin=2em,
columns=fullflexible,
keepspaces=true,
classoffset=2,
morekeywords={self,Net,super,__init__},
keywordstyle={\color{codeblue}\bfseries},
classoffset=3,
morekeywords={class, def, return},
keywordstyle={\color{codedarkpurple}\bfseries},
stepnumber=1,
numbers=none,
captionpos=b,
showspaces=false,
showstringspaces=false,
morestring=[b]",
frame=none,
}

\usepackage[capitalize]{cleveref}
\crefname{section}{Sec.}{Secs.}
\Crefname{figure}{Fig.}{Figs.}
\Crefname{section}{Section}{Sections}
\Crefname{table}{Table}{Tables}
\crefname{table}{Tab.}{Tabs.}

\def\cvprPaperID{9361} 
\def\confName{CVPR}
\def\confYear{2022}

\begin{document}
\title{Learning to Solve Hard Minimal Problems\thanks{This work was partially supported by projects: EU RDF IMPACT No.~CZ.02.1.01/0.0/0.0/15 003/0000468, EU H2020 ARtwin No.~856994.
The research of AL is partially supported by NSF DMS-2001267. 
TD acknowledges support from an NSF Mathematical Sciences Postdoctoral Research Fellowship (DMS-2103310.)
}}
\author{Petr Hruby\\
ETH Zürich\\
Department of Computer Science\\
{\tt\small petr.hruby@inf.ethz.ch}
\and
Timothy Duff\\
University of Washington\\
Department of Mathematics\\
{\tt\small timduff@uw.edu}

\and
Anton Leykin\\
Georgia Institute of Technology\\
School of Mathematics\\
{\tt\small leykin@math.gatech.edu}

\and
Tomas Pajdla\\
Czech Technical University in Prague\\
Czech Institute of Informatics, Robotics and Cybernetics\\
{\tt\small pajdla@cvut.cz}

}
\maketitle

\begin{abstract}
   \noindent We present an approach to solving hard geometric optimization problems in the RANSAC framework. The hard minimal problems arise from relaxing the original geometric optimization problem into a minimal problem with many spurious solutions. Our approach avoids computing large numbers of spurious solutions. We design a learning strategy for selecting a starting problem-solution pair that can be numerically continued to the problem and the solution of interest. We demonstrate our approach by developing a RANSAC solver for the problem of computing the relative pose of three calibrated cameras, via a minimal relaxation using four points in each view. On average, we can solve a single problem in under 70 $\mu s.$ We also benchmark and study our engineering choices on the very familiar problem of computing the relative pose of two calibrated cameras, via the minimal case of five points in two views.
\end{abstract}

\section{Introduction}
Minimal problems arise from geometrical problems in 3D reconstruction~\cite{Snavely-SIGGRAPH-2006,snavely2008modeling,schoenberger2016sfm}, image matching~\cite{rocco2018neighbourhood}, visual oodometry,and localization~\cite{Nister04visualodometry,Alismail-odometry,Sattler-PAMI-2017,taira2018inloc}. Many geometrical problems have been successfully formulated and solved as minimal problems~\cite{Nister-5pt-PAMI-2004,Stewenius-ISPRS-2006,kukelova2008automatic,Byrod-ECCV-2008,DBLP:conf/cvpr/RamalingamS08,Elqursh-CVPR-2011,mirzaei2011optimal,DBLP:conf/eccv/KneipSP12,Hartley-PAMI-2012,kuang-astrom-2espc2-13,Kuang-ICCV-2013,saurer2015minimal,ventura2015efficient,DBLP:conf/eccv/CamposecoSP16,SalaunMM-ECCV-2016,AgarwalLST17,Barath-CVPR-2017,Barath-CVPR-2018,Barath-TIP-2018,Miraldo-ECCV-2018}. Technically, minimal problems are systems of polynomial equations which depend on the input data and have a finite number of solutions. 

\subsection{Motivation}
\begin{figure}[t]
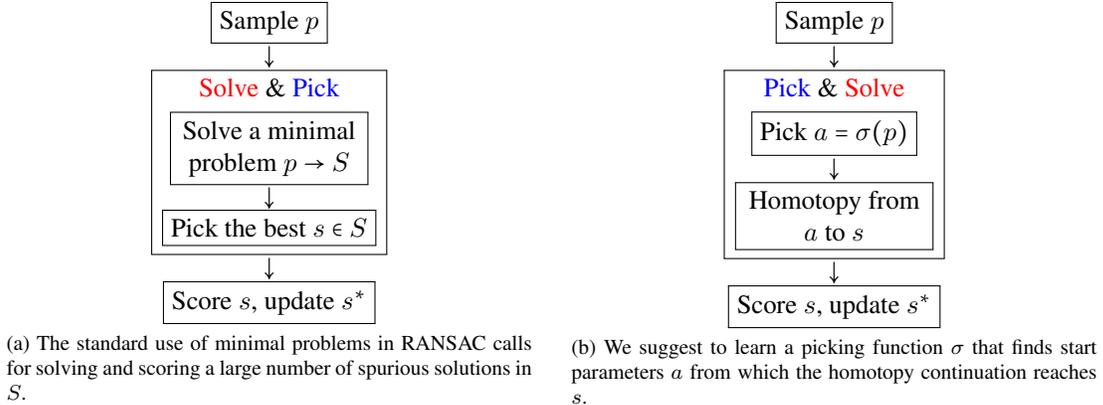

\centering
\begin{subfigure}[t]{0.4\linewidth}
\centering
\fbox{Sample $p$}\\
$\downarrow$\\
\fbox{\parbox{2.9cm}{\centering {\color{red} Solve} \& {\color{blue} Pick} \\[1ex]
\fbox{\parbox{2.4cm}{\centering Solve a minimal problem $p \rightarrow S$}}\\
$\downarrow$\\
\fbox{Pick the best $s \in S$}
}}\\
$\downarrow$\\
\fbox{Score $s$, update $s^*$}
\caption{The standard use of minimal problems in RANSAC calls for solving and scoring a large number of spurious solutions in $S$.}
\end{subfigure}
\hspace*{1em}
\begin{subfigure}[t]{0.4\linewidth}
\centering
\fbox{Sample $p$}\\
$\downarrow$\\
\fbox{\parbox{2.7cm}{\centering {\color{blue} Pick} \& {\color{red} Solve} \\[1ex]
\fbox{Pick $a = \sigma(p)$}\\
$\downarrow$\\
\fbox{\parbox{2.4cm}{\centering Homotopy from $a$ to $s$}}\\
}}\\
$\downarrow$\\
\fbox{Score $s$, update $s^*$}
\caption{We suggest to learn a picking function $\sigma$ that finds start parameters $a$  from which the homotopy continuation reaches $s$.}
\end{subfigure}
\caption{The inner RANSAC loop finds the best solution for a data sample $p$. This is very expensive when a minimal problem has many spurious solutions. Our efficient homotopy continuation combined with machine learning avoids solving for the spurious solutions. Thus, using minimal problems in RANSAC becomes effectively independent from the number of spurious solutions.}
\label{fig:teaser}
\end{figure}
Many geometrical problems are optimization problems that have only one optimal solution. Minimal problems, however, often have many additional spurious solutions. The optimal solution is typically real, satisfies inequality constraints, and fits well all data. Such constraints, however,  can not be used by methods of nonlinear algebra~\cite{Cox-UAG-1998,Sturmfels-CBMS-2002}
which have no ability to bypass finding (or incurring the cost of finding) all solutions of polynomial systems. 

RANSAC~\cite{Fischler-Bolles-ACM-1981,Raguram-USAC-PAMI-2013} approximates the optimal solution to a geometrical problem by computing candidate solutions from data samples and picking a solution with maximal data support.
This is done by iterating over the samples in an outer loop and over the solutions of a minimal problem for each sample in an inner loop. To find a single solution for a data sample in the inner loop, the state-of-the-art ``solve \& pick'' approach first computes all solutions of a minimal problem and then picks the optimal solutions by removing nonreal solutions, using inequalities, and evaluating the support. Optimization in the inner loop may be very costly when there are many spurious solutions to the minimal problem. Fig.~\ref{fig:teaser} compares the standard ``solve \& pick'' approach with our ``pick \& solve'' approach that learns, for a given data sample, how to first pick a promising starting point and then (ideally) continue it to a meaningful solution. 

Recent results~\cite{PLMP, PL1P} show that there are many minimal problems in multiview geometry with many spurious solutions which the state-of-the-art polynomial solvers cannot solve efficiently. \footnote{See Sec.~\ref{supp:important-mps} in the SM for more about these problems.}

\subsection{Contribution}
We present a method for combining optimized homotopy continuation (HC) with machine learning to avoid solving for spurious solutions. The main idea is to learn a single starting point for a real HC path that has a good chance to reach a good solution of the original geometrical problem. 
\hide{By avoiding spurious solutions, we can address the efficient solving of geometrical problems, which could never be solved efficiently by the state-of-the-art symbolic-numeric solvers for the corresponding minimal problems.} 

To demonstrate our method on a hard problem, we develop an efficient solver for the ``Scranton'' minimal problem obtained by relaxing the overconstrained problem of four points in three views (4pt)~\cite{Nister-IJCV-2006}. 
We train a model that predicts a starting problem for a single path real HC method to find a good solution. Our solver is implemented efficiently in C++ and evaluated on the state-of-the-art data in computer vision. It successfully solves about 26.3\% of inputs in 16.3$\mu s$, Tab.~\ref{tab:eval_mlp}. In~Sec.\ref{sec:experiments} we show that when used in RANSAC, about 4 samples 
suffice on average to obtain a valid candidate of camera geometry in 61.6$\mu s$. 
\arxiv{No such efficient solver has been known for this problem before. 
The best-known runtime for a very carefully designed approximation of the problem, reported in~\cite{Nister-IJCV-2006}, was on the order of milliseconds. We thus achieve more than ten times speedup compared to~\cite{Nister-IJCV-2006}. Most importantly, our approach is general and opens the door to solving other hard minimal problems, e.g., from~\cite{PLMP,PL1P}.}

We benchmark (Sec.~\ref{sec:experiments}) our approach on the classical 5-point problem (5pt)~\cite{Nister-5pt-PAMI-2004} using  standard benchmarks~\cite{ransac_tutorial}. We show that for the 5pt problem, we can solve 29.0$\%$ of inputs in about 7.6$\mu s$, Tab.~\ref{tab:eval_mlp}. Thus, in RANSAC, we can solve it in average in 26.1$\mu s$.  

\hide{It demonstrates that our approach is capable of achieving practical runtimes even for relatively simple problems, for which we have highly optimized symbolic-numeric solvers~\cite{larsson2017making}. This shows that current symbolic-numeric solvers are best for solving simple minimal problems with a small number of solutions. Our method brings the best solving times for hard minimal problems with many spurious solutions.}

Our approach is general. 
\arxiv{It can be applied even in some cases where the number of spurious solutions is not finite. For instance, our depth formulation of the Scranton problem has an infinite family of solutions where some depths may be zero.
Additional polynomial constraints, which do not need to be explicitly enforced, reduce the number of potential solutions to $272$---see~\ref{sec:Scranton-formulation-details}.
Thus, by exploiting the ``locality'' of HC methods, we can guarantee that when starting from a good starting point, we can ignore other spurious solutions with no additional computational cost. 

}


It is important to highlight that, unlike the current symbolic-numeric solvers, our method is coupled with the rest of SfM pipeline, i.e., it uses the real data distribution in a particular vision problem at hand.  

\hide{The additional strength of our method lies in this coupling. For instance, while losing in a direct comparison of runtimes on {\em one instance} of a 5-point problem to the state-of-the-art (agnostic) solvers, our approach performs on par with these if used in the SfM pipeline involving RANSAC.}

\subsection{Previous work}
\label{sec:prev}

The state-of-the-art approach to solving polynomial systems in computer vision is based on symbolic-numeric solvers, which combine elimination (by Gr\"obner bases~\cite{Stewenius-ISPRS-2006,kukelova2008automatic,larsson2017efficient} or resultants~\cite{DBLP:journals/corr/abs-1201-5810,DBLP:conf/iccv/Heikkila17,DBLP:conf/cvpr/BhayaniKH20}) with eigenvector computation~\cite{Sturmfels-CBMS-2002} to find all complex solutions. Currently, symbolic-numeric solvers~\cite{kukelova2008automatic,larsson2017efficient,Larsson-Saturated-ICCV-2017,larsson2017making} provide efficient and stable results for minimal problems with as many as 64 complex solutions~\cite{Larsson-CVPR-2018,DBLP:conf/cvpr/BhayaniKH20}. However, these solvers mostly fail to deliver practical results for hard computer vision problems with many solutions. Symbolic/numeric solvers involve two hard computational tasks. First, in the symbolic part, large matrices are constructed and triangularized by the Gauss-Jordan elimination. Secondly, in the numeric part, the eigenvectors of $n \times n$ matrices, where $n$ is the number of all complex solutions, are computed. Both steps are prohibitive for generic polynomial systems with many spurious solutions. 
Methods which deal with real solutions only, e.g.\ Sturm sequences~\cite{Nister-5pt-PAMI-2004}, also generally require expensive manipulations (reduction to a univariate polynomial) and may still need to consider many spurious real solutions.
\hide{The numeric part can be sometimes simplified by producing a univariate polynomial in the symbolic part and then using bracketing, e.g.\ Sturm sequences~\cite{Nister-5pt-PAMI-2004}, to deal with the real solutions only. This may sometimes dramatically reduce the number of spurious solutions to deal with, but in general, there still may remain many spurious real solutions.}
\hide{Another problem is that constructing univariate polynomials adds additional computation to the constructions used in classical symbolic/numeric solvers~\cite{}.}

Global HC methods give an alternative, well-studied approach~\cite{DBLP:journals/cca/Verschelde10,DBLP:books/daglib/0032895,Duff-Monodromy,DBLP:conf/icms/BreidingT18} to finding \emph{all complex solutions} of polynomial systems. 
Off-the-shelf HC solvers~\cite{DBLP:journals/cca/Verschelde10,DBLP:books/daglib/0032895,Duff-Monodromy,DBLP:conf/icms/BreidingT18} have been proven useful for studying the structure of minimal problems~\cite{Holt-PAMI-1995,QuanTM2006,AholtO14,Kileel-MPCTV-2016}. 
However, the off-the-shelf solvers are much slower ($10^3-10^5$ times) than current symbolic-numeric solvers. 
For instance, our experiments, \cref{tab:ablation-hc-study}, show that solving~\cite{Nister-5pt-PAMI-2004} with complex homotopy continuation in Macaulay2~\cite{leykin2011numerical} takes about $10^5 \mu s$ compared to $5\, \mu s$ when~\cite{PoseLib} implementation of~\cite{Nister-5pt-PAMI-2004}  is used.
\hide{General homotopy continuation methods use numerical computation to track all solutions of a starting (easy) minimal problem to the target (hard) minimal problem. They are guaranteed to find the solutions of the target system only when tracking all solutions in the complex numbers~\cite{DBLP:books/daglib/0032895}. Thus, general homotopy continuation also fails to to deliver practical results for hard computer vision problems with many complex solutions.}

The previous work~\cite{TRPLP} closest to this work addresses the problem of speeding up minimal HC solvers. \arxiv{This paper took notable steps towards the practical use of homotopy continuation for minimal problem solving.} In that work, an off-the-shelf HC implementation~\cite{leykin2011numerical} has been optimized and efficiently implemented in modern C++. Two hard problems in trifocal geometry have been solved in about 660$m s$. Despite not providing practical solvers for RANSAC~\cite{Fischler-Bolles-ACM-1981,Raguram-USAC-PAMI-2013}, the results of~\cite{TRPLP} demonstrated that hard multiview problems involving 312 and 216 solutions can be solved in a stable way and thus could be useful for building practical structure from motion algorithms. 
\hide{One might hope that  additional optimization for a particular polynomial system coming from a particular vision application could bring the desired speed. However, it is not the case. To find all solutions in, e.g., 62$\mu s$, which we can achieve~\ref{sec:} for the problem~\cite{Nister-IJCV-2006,Kileel-MPCTV-2016,PL1P} with 272 solutions, one solution would have to be tracked under 62$\mu s$/272=0.23$\mu s$. This is currently not possible unless parallelized.\marginpar{I'd just delete this sentence... nobody is talking about parallelizing. We can have 272 threads computing each path in less than 61$\mu s$.}
}

Our paper considers a novel solver for a problem named ``Scranton"\footnote{The US telephone area code for Scranton PA is 272.}, which is a minimal relaxation of the overconstrained 4pt problem~\cite{Nister-IJCV-2006}. 
Previous work formulated Scranton using camera parameters and found that it has 272 complex solutions~\cite{Kileel-MPCTV-2016,PL1P}.
We consider an alternative formulation in terms of 3D point depths, analogous to~\cite{QuanTM2006}, which has 272 potentially meaningful solutions.
\hide{\cite{QuanTM2006} has various numbers of solutions depending on the dehomogenization of the problem~\cite{QuanTM2006}. We use the depth formulation with the simplest dehomogenization setting the first depth to one. That brings additional spurious solutions over the formulation with 272 solutions, but we can do it since the additional solutions do not affect our solving method which tracks a single path only. No solver for Scranton has been published before.} The original 4pt problem was solved by numerical search in~\cite{Nister-IJCV-2006}, with about $1 ms$ runtime. 
A depth-formulated 4pt problem was also studied in~\cite{QuanTM2006},  showing that the overconstrained problem with exact input has a unique solution. 
Their exact solution does not apply to problems with noisy data.
\hide{They verified their results by solving an exact system in a prime field by computing its Gr\"obner basis, which cannot be used to solve practical noisy problems.}

We also study the classical, well-understood 5-point problem (5pt) of computing the relative pose of two calibrated cameras~\cite{Nister-5pt-PAMI-2004}. 
Unlike Scranton, this problem has many practical solutions~\cite{Nister-5pt-PAMI-2004,Hartley-PAMI-2012,Kukelova-PolyEig-PAMI-2012,opencv_library,schoenberger2016sfm}. Currently, the most efficient symbolic-numeric solver~\cite{larsson2017making} of the 5pt problem solves for up to 10 essential matrices in about $5 \mu s$. Although the 5pt problem is not as hard as Scranton in terms of the number of spurious solutions, it does provide an important testing ground for us.

\section{Our approach}\label{sec:Approach}
Here we present our approach to solving hard minimal problems. We shall use HC methods to track one real solution of a start problem to obtain one real solution of the target problem. We shall design an algorithm such that this one solution we obtain is a \emph{meaningful} solution with sufficient probability.  

\subsection{Problem-solution manifold}
\label{sec:Problem-solution manifold}
We operate in the \emph{problem-solution manifold} $M$ of \emph{problem-solution (p-s) pairs} $(p,s)$, where $p$ is a problem and $s$ is a solution of $p$. Problem $p$ belongs to a \emph{real} vector space $P$. Solution $s$ comes from a \emph{real} vector space of solutions. 
The projection $\pi: M\to P$ is defined by $(p,s)\mapsto p$. The preimage $\pi^{-1}(p) \in M$ contains all p-s pairs that correspond to a particular problem~$p$.

\begin{figure}[t]
    \centering
    \begin{subfigure}[t]{0.4\linewidth}
        \centering
        \begin{overpic}[width=0.9\linewidth
        ]{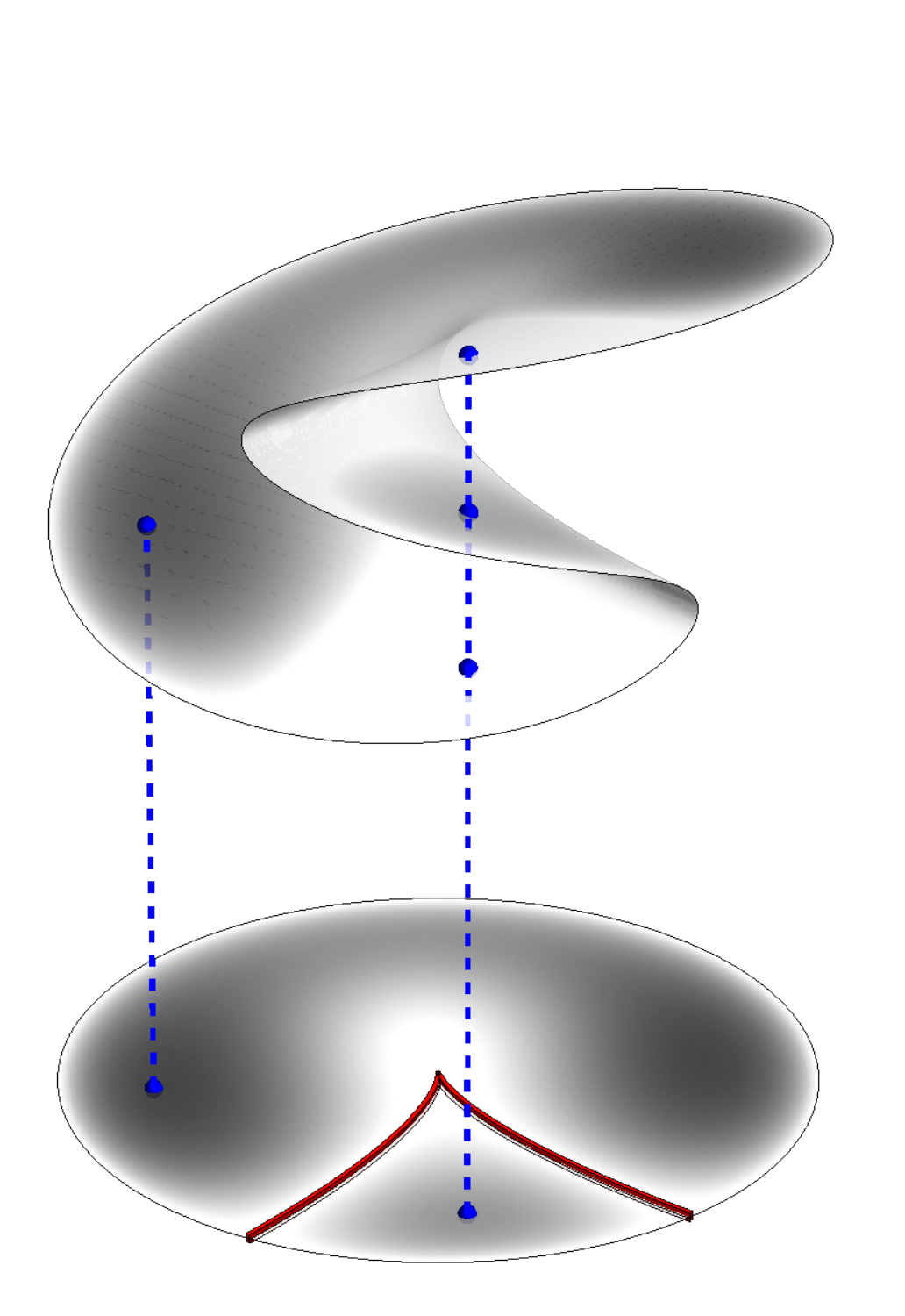}
        \put (53,43) {$\def\arraystretch{1.5}\displaystyle \begin{array}{c}
             M\\
             
             \phantom{\pi}\Big\downarrow \pi\\
             
             P 
        \end{array}$}
        \end{overpic}
        \caption{}
        \label{fig:M-and-mu}
    \end{subfigure}%
    ~ 
    \begin{subfigure}[t]{0.58\linewidth}
        \centering
        \includegraphics[width=0.9\linewidth]{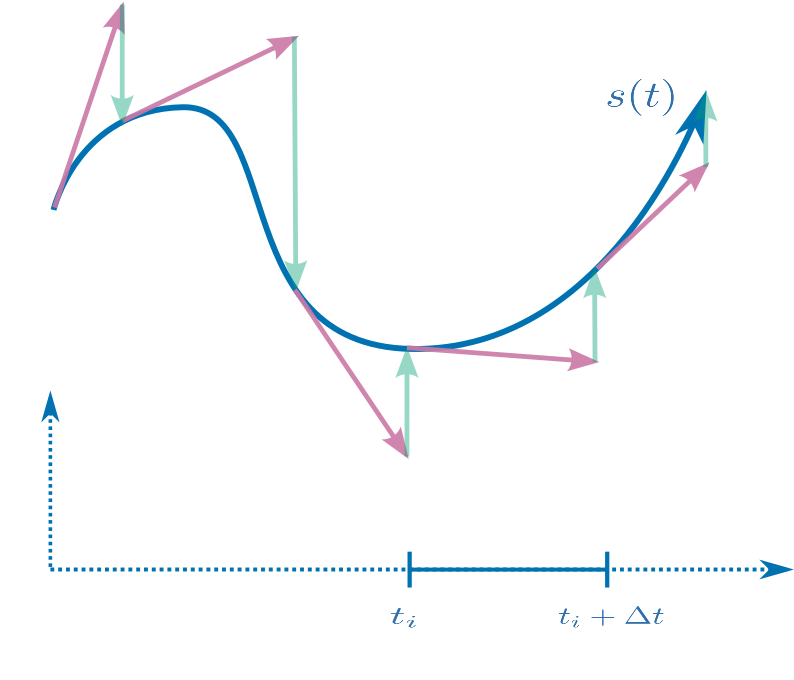}
        \caption{}
        \label{fig:hc}
    \end{subfigure}
        \caption{(a) Problem-solution manifold $M$ projected to the problem space $P$. (b) Numerical HC method.}
\end{figure}

\begin{example}
\label{ex:cubic}
To illustrate the introduced concepts, let us look at one equation $x^3+ax+b=0$ in one unknown $x$ with two parameters $a$ and $b$. Here a problem $p=(a,b)$ has either one or three real solutions depending on whether the discriminant $D = 4a^3 + 27b^2$ is positive or negative; see the corresponding problem-solution manifold $M$ in Fig.~\ref{fig:M-and-mu}. 

To be precise, the equation defines an algebraic variety, which is guaranteed to be a smooth manifold when points above the discriminant locus $D=0$ are removed. 
\end{example}

See SM Sec.~\ref{sec:PS-M-Examples} for the detailed examples of setting up problem-solution manifolds for the 5pt problem and Scranton, a minimal relaxation for the 4pt problem. Below is a condensed version of the 5pt problem.

\begin{example}
\label{ex:5p-SM-condensed}
Consider the classical 5pt problem of computing the relative pose of two calibrated cameras which view five world points where the scale is fixed such that the first 3D point lies in the first image plane.

Denote by $x$ the images of 5 points in 2 views: i.e., $x$ is a point in $P = \RR^{20}$. Assume the points are in front of both cameras: i.e.,  their depths $\lambda_{i,j}, (i=1,\dots,5; j=1,2)$ are all positive.
Our assumptions imply that $\lambda_{1,1}=1$. 
The unknown depths vector in the solution space $\RR^9$ determines the relative pose. Therefore, the problem-solution manifold $M$ is contained in the ambient space $\RR^{20} \times \RR^{9}$. Equations vanishing on $M$ are given in Sec.~\ref{sec:5pt}.
\end{example}

\subsection{Probability distribution on $M$}
\label{sec:prob-dist}
We introduce a probability density $\mu$ on the problem-solution manifold $M$ that gives the distribution of real-world problem-solution pairs. In \Cref{fig:M-and-mu}, we give an example of $\mu$ depicted with shades of gray (darker is bigger) on the manifold $M$. Note that the density is such that any problem with a set $S$ of three solutions has only one $s\in S$ with $\mu(s)>0$. We shall (implicitly) operate under the following assumption:

\emph{An input problem $p$ is likely to have one meaningful solution that is dominant, i.e., occurs much more frequently in the real data than other meaningful solutions.}

In many problems (e.g., the one in \Cref{sec:4pt4vFormulation}) the number of meaningful solutions is guaranteed to be exactly one generically. The distribution $\mu$ is hard to model; in what follows it is represented by training data.

\subsection{Pick \& solve vs.\ solve \& pick}

A typical local iterative method (e.g.\ Newton's method, gradient descent, etc.) would attempt solving a problem $p$ by obtaining an initial approximate solution $s_0$ with a hope that it is not too far away from an actual solution and then producing a sequence of its refinements $s_0, s_1, s_2, \dots$ until either a desired quality is reached or some termination-with-failure criterion is satisfied.  

Our homotopy approach is a generalization of such local methods. In a nutshell, given a problem $p$,
\begin{enumerate}[itemsep=1pt,parsep=1pt]
    \item we select a suitable start problem-solution pair $(p_0,s_0)\in M$ for $p$,\label{step:start}
    \item we choose a path $p_{0}\leadsto p$ in the problem space $P$, leading from $p_0$ to $p$,\label{step:path}
    \item we track the path $(p_0,s_0) \leadsto (p,s)$ to obtain the target solution $s$ of $p$.\label{step:track}
\end{enumerate}

Selecting a start pair $(p_0,s_0)$ is the key ingredient of our approach. Given a real HC method, one can aim to construct the selection strategy $\sigma(p) = (p_0,s_0)$ in two steps. First, one finds a small set of \emph{anchors} $A \subset M$, such that it is possible to reach (cover) a significant part of $M$ from $A$ by the real homotopy continuation. Secondly, one learns a selection strategy  $\sigma$ such that starting from $(p_0,s_0) \in A$, the meaningful problem-solution pair is reached with sufficiently high frequency to make RANSAC work.

Intuitively, one may perceive a minimal solver  employed in RANSAC as an arrow $p\to S$ in the following diagram:   
$$
\begin{tikzcd}[column sep=16ex]
p \ar[r,"\text{solve (minimal problem)}"] 
\ar[rd,"\text{pick (anchor)}",bend right=20]& S=\pi^{-1}(p)\ar[r,"\text{pick ($s \in S$)}"]& s \\
&a=(p_0,s_0)\ar[ru,"\text{solve (homotopy)}",bend right=20] &
\end{tikzcd}
$$
where an instance $p$ of a minimal problem is ``solved'' (all solutions in $S$ are found.) Then RANSAC ``picks'' at most one solution, $S \rightarrow s$, a candidate that maximizes the number of inliers.

Looking for a shortcut that would allow us to go directly $p\rightarrow s$, we reverse this flow: first ``picking'' an anchor and then tracking \emph{one} HC path to ``solve''. 


\subsection{Structure of our solvers}\label{sec:summary}
\hide{We might want to create a section, which would contain the specific description of our solver. This section would serve as a bridge between the general principles above and the specific implementations of the parts of the solver in the sections below. This section would contain a description of the algorithm together with links to the sections where the subproblems are solved.}

Our solvers for both 5pt and Scranton minimal problems have a common structure, consisting of an offline training stage and an online evaluation stage. 
Offline computations may be resource-intensive; however, the online stage must be very efficient to achieve practical sub-millisecond run times. The offline stage consists of:
\begin{enumerate}[itemsep=1pt,parsep=1pt]
    \item Sampling data $D,$ according to $\mu ,$ representing the (preprocessed) problem-solution manifold $M$ (Sec~\ref{sec:A0}).
    \item Covering a sufficient fraction of the data with anchors $A \subset D$
    (Sec.~\ref{sec:anchors}).
    \item Learning a model $\sigma$ which selects a starting ps-pair $(p_0,s_0) \in A$ for any given problem $p$ (Sec.~\ref{sec:start-p-s}).
\end{enumerate}
The online stage consists of:
\begin{enumerate}[itemsep=1pt,parsep=1pt]
    \item Preprocessing the input $p$ to reduce its variability (Sec.~\ref{sec:preprocess}).
    \item \hide{\tim{Technically, the preprocessed problem is $v.$ Maybe this technical detail should be ignored here.}}
    Selecting a starting pair from $A$ as $(p_0, s_0) = \sigma(p)$.
    \item Constructing polynomial equations of $p$ (Sec.~\ref{sec:formulation}).
    \item Computing solution $s$ of $p$ by HC from $(p_0, s_0)$ (Sec.~\ref{sec:hc_solving}).
    \item Recover solution $s$ of the original problem $p$ (Sec.~\ref{sec:preprocess}).
\end{enumerate}
The next sections describe these steps in detail.

\section{Sampling data representing $M$ and $\mu$}
\label{sec:A0}

The offline stages of our solvers begin by sampling the data $D$ given by 3D models of various realistic objects. We use models from ETH 3D Dataset to represent $\mu$.

A 3D model consists of 3D points $X$, cameras $C$, and relation $I \subset X \times C$ encoding observations ($(X_m, C_i) \in I$ iff $C_i$ observes $X_m$). For Scranton, we may sample a single p-s pair $(p,s) \in M$ as follows:
\begin{enumerate}[itemsep=1pt,parsep=1pt]
    \item Select $3$ cameras $C_i, C_j, C_k \in C$
    \item Select $4$ points $X_l, X_m, X_n, X_o \in X:$ $(X_a, C_b)\in~I \ \forall a \in \{l,m,n,o\}, b \in \{i,j,k\}$
    \item Project the points to the cameras to get $12$ 2D points $x_{a,b}$, concatenate them to $24$-dim vector $p \in P$
    \item Get the depths $\lambda_{a,b}$ of the points in the cameras, concatenate them to $12$-dim vector $s \in S$
\end{enumerate}
Sampling for the 5pt problem is similar; in step 1 we select $2$ cameras, and in step 2 we select $5$ points.

\section{Selecting anchors $A$}
\label{sec:anchors}
\label{sec:generate_anchors}
\renewcommand{\tabcolsep}{3pt}
\begin{table}
    \centering
    
    \begin{tabular}{|l||c|c|c|c||c|c|c|c|}
      \hline
      & \multicolumn{4}{c||}{5pt problem} & \multicolumn{4}{c|}{Scranton}\\      
      \hline
      n & 1k & 4k & 10k & 40k & 1k & 4k & 10k & 40k\\
      \hline
      50 \% & 8 & 9 & 8 & 8 & 18 & 18 & 17 & 16 \\
      75 \% & 25 & 28 & 27 & 26 & 47 & 51 & 50 & 50 \\
      90 \% & 59 & 70 & 70 & 70 & 92 & 112 & 120 & 134 \\
      95 \% & \textit{90} & 109 & 115 & 124 & \textit{126} & 168 & 191 & 233 \\
      100 \% & \textit{140} & \textit{235} & \textit{334} & \textit{585} & \textit{176} & \textit{335} & \textit{507} & \textit{1205} \\
      \hline
  \end{tabular}

    \caption{Number of anchors obtained according to Sec.~\ref{sec:anchors} which are needed to cover $50 \%$, $75 \%$, $90 \%$, $95 \%$, and $100 \%$ of $n$ problem-solution pairs. Different values of $n$ are considered. A problem-solution pair is covered by the set of anchors if there is at least one anchor from which the problem-solution pair can be correctly tracked. A track is considered correct if the Euclidean distance from the obtained solution to the ground-truth solution is less than $10^{-5}$.}
    \label{tab:anchors}
\end{table}
We now describe how the starting p-s pairs are obtained. Our goal is to find a small set $A$ of starting p-s pairs (the set of anchors), from which a high portion of p-s pairs from a given distribution can be tracked by HC. 

\begin{figure*}[t]
    \centering
    \includegraphics[width=\linewidth]{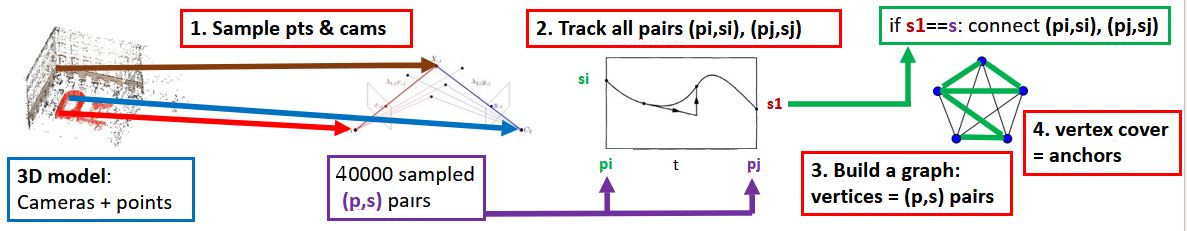}
    \caption{Illustration of generating anchors. A minimal sample of cameras and points is sampled from an existing 3D model. Then, the sampled geometry is converted to the problem-solution pairs. A graph is built whose nodes are the sampled problem-solution pairs. HC is tracked from every p-s pair to every other p-s pair. If the obtained solution is equal to the sampled solution, the p-s pairs are connected with an edge. The selected anchors are a vertex cover of the graph obtained with a greedy algorithm.}
    \label{fig:anchors}
\end{figure*}

If we limit ourselves to a finite set of p-s pairs and the anchors are selected from the same set, the optimal procedure for the anchor selection consists of building a graph with p-s pairs as vertices. The nodes $(p_i,s_i)$, $(p_j, s_j)$ are connected with an edge, if the correct solution $s_j$ can be obtained by tracking HC from $(p_i, s_i)$ to $p_j$. A set of anchors covering all problems in this graph is called a \emph{dominating set.}
Since computing a minimum-size dominating set is NP-hard,  we replace it with a greedy proxy, which is known to perform well\footnote{The proposed method is illustrated in SM.~Fig.~\ref{fig:anchors}}.

To show that the selected anchors generalize well to p-s pairs from other scenes, we consider $40000$ anchors generated from \textit{office} and \textit{terrains}. For testing data, we generated p-s pairs from the models \textit{delivery\_area}, and \textit{facade}.

We have tracked the solutions with HC starting from each of the anchors. The portion of problems correctly tracked from any of the anchors is shown in Tab.~\ref{tab:coverage}. The resulting percentage is equivalent to using an oracle that always finds the best anchor to start from.

Next, we show that if the number of vertices in the graph is sufficiently high, a reasonable portion of different p-s pairs from the same scene can be solved by HC starting from one of the anchors $A$ we generate. 

We have generated $n$ problem-solution pairs from models \textit{office} and \textit{terrains}\footnote{https://www.eth3d.net/datasets}. Out of these problem-solution pairs, we have selected anchors which cover $50 \%$, $75 \%$, $90 \%$, $95 \%$, and $100 \%$ of data. Tab.~\ref{tab:anchors} shows the number of anchors which cover the data for different values of $n$ for both problems. The number of anchors for $50 \%$ and $75 \%$ saturates when $n=10000$. Therefore, we may assume that these anchors will cover a significant portion of problems from the given distribution. Tab.~\ref{tab:anchor_sources} shows a comparison of different sources of anchors. The combination of models \textit{office} and \textit{terrains} gives the best generalizability out of all considered sources.

\begin{table}
\begin{center}
  \begin{tabular}{|l|c||l|c|}
    \hline
    Source    & $\alpha$ [\%] & Source    & $\alpha$ [\%]\\ \hline \hline
    Courtyard & 78.1 & Relief 2 & 77.0 \\ \hline
    Office & 81.1 & Off. + Terr. & \textbf{82.2} \\ \hline
    Terrains & 79.0 & Off. + Rel. & 80.7 \\ \hline
    Playground & 75.4 & O + T + P + R2 & 79.7 \\ \hline
  \end{tabular}
\end{center}
\caption{Study of sources for anchor selection. Rows correspond to different models or combinations of models, from which the anchors are generated. For each source of anchors, we measure the percentage $\alpha$ of testing p-s problems (generated from models \textit{delivery\_area}, \textit{electro}, \textit{facade}, \textit{kicker}, \textit{meadow}, \textit{pipes}) that can be reached from any of the anchors generated from the given source. Anchor sets of 100 anchors are considered for every source. }
\label{tab:anchor_sources}
\end{table}
\section{Learning $\sigma$ to select the starting p-s pair}
\label{sec:start-p-s}

\begin{table}[t]
    \centering
    \begin{tabular}{|l||c|c|c|c|c|}
    \hline
    & \multicolumn{5}{c|}{5 pt problem coverage [\%]}\\
    \hline
    \# anchors & 8 & 26 & 70 & 124 & 585 \\
    \hline
    delivery\_area & 43.3 & 73.5 & 86.8 & 91.4 & 96.5 \\
    facade & 51.3 & 74.9 & 88.6 & 92.9 & 97.0\\
    \hline
    \hline
    & \multicolumn{5}{c|}{4 pt problem coverage [\%]}\\
    \hline
    \# anchors & 16 & 50 & 134 & 233 & 1205 \\
    \hline
    delivery\_area & 47.4 & 73.6 & 88.2 & 92.9 & 96.6 \\
    facade & 42.1 & 71.1 & 87.9 & 92.9 & 96.9\\
    \hline
    \end{tabular}
    \caption{Percentage of testing problem-solution pairs which are solvable by the anchors generated from model \textit{Office} and \textit{Terrains}. The anchors are taken from Tab.~\ref{tab:anchors} for $n = 40000$. The problem-solution pair is considered solvable by the anchors, if the correct solution can be obtained by HC starting in any of the anchors. The solution is considered correct if the Euclidean distance from the obtained solution to the ground-truth solution is less than $10^{-5}$. This is equivalent to using an oracle classifier that always finds the best anchor to start from.}
    \label{tab:coverage}
\end{table}

We formulate the problem of finding the best starting p-s pair as a classification task. Our method relies on a classifier $\sigma $, which for a sample problem $p \in P$ assigns a label from $\sigma (p) \in A \cup \{TRASH\}$, where $A$ is the anchor set generated in Sec.~\ref{sec:generate_anchors}. 
The label $ TRASH$ is included for cases where no problem in $A$ covers $p.$

Our goal is to minimize the \emph{effective time} $\epsilon_t = \mu_t / \rho $ of the solver, where $\mu_t$ is the total time\footnote{preprocessing, anchor selection, tracking, and RANSAC scoring} and $\rho$ is the success rate. Therefore, we must be able to classify $p$ very fast, on the order of $10 \mu s$ for a subsequent HC path $\sigma (p) \leadsto (p, s)$  $\sigma (p) \ne \{TRASH\}.)$
Now, we are going to describe the classifier and its training.

For both problems, we use a Multi-Layer Perceptron (MLP) with $6$ hidden layers of $100$ neurons with bias.\footnote{See SM Tab.~\ref{tab:MLP-size-study} for a comparison with MLPs with different sizes.} The input layer has size $\dim P$, and the output layer has size $|A| + 1$. We use the PReLU activation function. During training, we use the dropout before the last layer to prevent overfitting. The classification time of the MLP is about $8 \mu s$ for both 5pt and 4pt problems.

The input to the MLP is a normalized (Sec.~\ref{sec:preprocess}) problem $p \in P$. The output is a vector of $|A| + 1$ numbers, which give the score for every starting p-s pair $(p_0, s_0)$, as well as for $TRASH$. If the score of $TRASH$ is higher than the scores of all anchors, we skip the sample. Otherwise, we track from the p-s pair with the highest score.

During training, we normalize the output of the MLP with a softmax layer, and we use a cross-entropy loss, which is a standard method for training classifiers. We use the SGD optimizer, which gives us better results than other optimizers, such as Adam. We generate training and testing data data according to Sec.~\ref{sec:train_data}, and train the MLP by minimizing the loss on the training data for 80 epochs. Then, we select the parameters which maximize the success rate on the validation data. The evaluation of the classifier is shown in Sec.~\ref{sec:evaluation}.


\subsection{Training data generation}\label{sec:train_data}
\begin{figure*}[t]
    \centering
    \includegraphics[width=\linewidth]{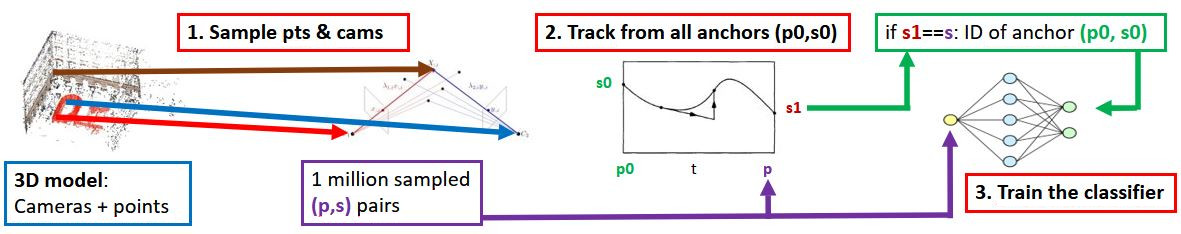}
    \caption{Illustration of generating training data and training the classifier. A minimal sample of cameras and points is sampled from an existing 3D model. Then, the sampled geometry is converted to the problem-solution pairs. Every generated problem-solution pair $(p,s)$ is tracked from every anchor $(p_0, s_0)$. If the correct solution is obtained, problem $p$ is added to the training data, whereby the ID of the anchor $(p_0, s_0)$ is used as the expected label. Then, the MLP is trained on the generated training data.}
    \label{fig:train_data}
\end{figure*}

We use the training and testing data generated from ETH 3D Dataset. Testing data is generated from models \textit{delivery\_area} and \textit{facade}, training data from 23 other sequences. First, we generated p-s pairs $(p, s)$ from the models according to Sec.~\ref{sec:A0}. Then, we normalized each problem $p$ (Sec.~\ref{sec:preprocess}), and tracked the solution to problem $p$ from each anchor $(\Breve{p}_a, \Breve{s}_a) \in A$. If the solution to $p$ obtained by HC starting in anchor $(\Breve{p}_a, \Breve{s}_a) \in A$ is equal to the expected solution $s$, then the ID $a$ of the anchor is assigned as the label of problem $p$. If solution $s$ cannot be reached from any anchor, the label of $p$ is $TRASH$. A problem may have multiple labels. We note that this procedure allows us, in principle, to generate an unlimited amount of training data\footnote{The procedure for generating training data is illustrated in Figure \ref{fig:train_data}.}.

In our experiments, we use about $1$ million training p-s pairs per model (23M in total) and $30000$ validation p-s pairs per model. 


\subsection{Classifier evaluation}\label{sec:evaluation}
Now, we are going to show the evaluation of the trained MLPs. During the evaluation, an anchor $(p_0, s_0)$ is selected by the classifier, and HC is tracked from $(p_0, s_0)$. \textbf{Success rate} is the percentage of test p-s pairs $(p,s)$ for which the correct solution $s$ is obtained by HC from $(p_0, s_0)$. The classification task is difficult because for some problems $p$, multiple geometrically meaningful solutions (all points in front of cameras, small ratios between the depths, and small baseline) exist. Therefore, we also consider MLP classifiers which return $m$ best anchors. Then, the classification is successful if the correct solution $s$ can be tracked from any of the selected anchors.

To show the benefits of our classifier, we also compare it with the following baselines:
\begin{enumerate}[itemsep=1pt,parsep=1pt]
    \item[B1] Start from every anchor in $A$.
    \item[B2] Start from the closest anchor $(p_0, s_0)$ in terms of Euclidean distance.
    \item[B3] Start from the closest anchor $(p_0, s_0)$ in terms of Mahalanobis distance.
\end{enumerate}

Note that the first baseline gives the upper bound on the success rate for a given anchor set $A$. The downside of this baseline is that HC paths from all $|A|$ anchors must be tracked. Success rate and total time for different classifiers are shown in Tab.~\ref{tab:eval_mlp}. The solution $s$ is considered correct if the squared Euclidean distance from the obtained solution to the ground-truth solution is less than $10^{-5}$.\\

\renewcommand{\tabcolsep}{2pt}

\begin{table}
\small
\begin{center}
  \begin{tabular}{|l||c|c|c||c|c|c|}
    \hline
     & \multicolumn{3}{c||}{5pt problem} & \multicolumn{3}{c|}{4pt problem }\\  \hline
     &\footnotesize $\rho$ [\%]& \footnotesize  $\mu_t$[$\mu s$]& \footnotesize $\epsilon_t$[$\mu s$]&\footnotesize  $\rho$[\%]& \footnotesize  $\mu_t$[$\mu s$]& \footnotesize $\epsilon_t$[$\mu s$]\\\hline \hline
    B1, $A_{50}$   &   47.3   & $\infty$ & $\infty$ &   44.2   & $\infty$ & $\infty$\\ \hline 
    B1, $A_{75}$   &   74.2   & $\infty$ & $\infty$ &   72.0   & $\infty$ & $\infty$\\ \hline 
    B1, $A_{90}$   &   87.7   & $\infty$ & $\infty$ &  87.9    & $\infty$ & $\infty$\\ \hline
    \hline
    B2, $A_{50}$   &   9.9   & 11.8 & 119.5 &   5.2   & 16.1 & 310.5\\ \hline 
    B2, $A_{75}$   &   9.0   & 12.4 & 137.8 &   4.9   & 16.7 & 340.9\\ \hline 
    B2, $A_{90}$   &   8.4   & 12.1 & 144.5 &   5.0   & 16.3 & 324.5\\ \hline
    B2, $A$ & 11.2 & 327.9 & 2927.7 & 9.8 & 150.1 & 1531.6    \\ \hline
    \hline
    B3, $A_{50}$   &   14.0   & 12.2 & 87.3 &   5.1   & 15.9 & 312.2\\ \hline 
    B3, $A_{75}$   &   13.4   & 12.8 & 95.3 &   4.8   & 17.0 & 352.7\\ \hline 
    B3, $A_{90}$   &   4.2   & 19.5 & 460.2 &   4.8   & 19.9 & 413.5\\ \hline
    \hline
    MLP, $A_{50}$  &    29.3  & 15.7 & 53.5 & 21.6 & 19.7 & 91.3 \\ \hline
    MLP, $A_{75}$  &   38.8   & 15.0  & 38.7  & 27.8 & 20.3 & 73.0 \\ \hline
    MLP, $A_{90}$  &   39.9   & 14.3 & 35.8 & 29.2 & 19.6 & 66.9 \\ \hline
    \hline
    \footnotesize MLP$_T$ $A_{50}$  &   17.0   & 4.6 & 26.9 & 9.1 & 8.9 & 96.8 \\ \hline
    \footnotesize MLP$_T$ $A_{75}$  &    29.0  & 7.6 & \textbf{26.1} & 19.0 & 13.5 & 71.1 \\ \hline
    \footnotesize MLP$_T$ $A_{90}$  &   36.8   & 10.8 & 29.3 & 26.3 & 16.2 & \textbf{61.6} \\ \hline
  \end{tabular}
\end{center}
\caption{Classifier evaluation. Rows correspond to start problem selection strategies. The anchors are extracted from datasets \textit{Office} and \textit{Terrains} (Tab.~\ref{tab:anchors}).  The strategies are evaluated on datasets \textit{Delivery\_area} and \textit{Facade}. $A_n$ denotes a set of anchors covering $n \%$ of the training datasets.  B1 tracks from all anchors in $A_n$. The success rate of B1 is equivalent to an ``Oracle'', which gives the best possible ``retrieval'' of the starting problem to reach the target problem, for a given set of anchors. ``B2, $A_n$'' selects the starting problem as the nearest anchor to the target problem (measured by the Euclidean distance in the space of normalized image points) from $A_n$. ``B3, $A_n$'' selects the starting problem as the nearest anchor to the target problem measured by the Mahalanobis distance. ``MLP+T, anchors $A_n$'' is our method selecting the starting problem as the one from $A_n$ with the highest score given by the MLP constructed in~Sec.\ref{sec:start-p-s}. Columns: $\rho$ is the success rate (recall) of retrieving a starting point from which the target problem can be reached, $\mu_t$ is the mean solving time, $\epsilon_t=100\,\mu_t/\rho$ is the mean effective solving time, i.e.\ the average time to obtain one correct solution. If multiple anchors are selected, the classification is considered successful if any of the tracks ends in a correct solution. A solution is considered correct if the squared Euclidean distance from the obtained solution to the ground-truth solution is less than $10^{-5}$. We measure total time needed to perform the classification, preprocessing of the input problem and HC from all selected anchors.}
\label{tab:eval_mlp}
\end{table}

\section{Homotopy continuation}
\label{sec:hc_solving}
We now recall the basic principles of HC methods \cite{DBLP:books/daglib/0014410,morgan2009solving,DBLP:books/daglib/0032895} in the framework of our work. Suppose we have a \emph{square system} of $n$ polynomial equations $f(p,s) = \left(f_1 (p,s), \ldots , f_n (p,s)\right)$ in $n$ unknowns $s=(s_1, \ldots , s_n)$ that vanish on our problem/solution manifold $M$. 

Our task is to numerically continue a known problem/solution pair $(p_0,s_0)\in M$ to a pair $(p,s) \in M$ for some problem of interest $p\in P.$ 
This may be accomplished by introducing a parameter homotopy $H(s,t) = f(p(t), s)$ where $p(t) : [0,1] \to P$ is some differentiable function with $p(0)=p_0$ and $p(1) = p.$ The goal is to compute a differentiable path $(p(t), s(t)):[0,1] \to M$ such that $s(0)=s_0$ and satisfying the implicit equation $H(p(t), s(t)) = 0$. Note that the homotopy $H$ depends on $p(t),$ and that many choices are possible.
We mainly consider {\bf Linear segment HC}: that is, we choose 
$p(t) = (1-t)\,p_0 + t\,p$. 

\definecolor{myblue}{RGB}{0,114,178}
\definecolor{mygreen}{RGB}{0,158,115}
\definecolor{myred}{RGB}{204,121,167}
\newcommand{\textmyblue}[1]{\textcolor{myblue}{#1}}
\newcommand{\textmygreen}[1]{\textcolor{mygreen}{#1}}
\newcommand{\textmyred}[1]{\textcolor{myred}{#1}}

In practice, we compute an approximation of the solution curve $s(t)$ by numerical \textmyred{predictor}/\textmygreen{corrector} methods, illustrated in~\cref{fig:hc}. 
In our \textmyred{predictor} step, the value $s(t_i)^*$ for a given $t_i \in [0,1)$ is known, and the value $s(t_i+\Delta t)$ for an adaptively-chosen stepsize $\Delta t$ is approximated using the standard fourth-order Runge-Kutta method. In the \textmygreen{corrector} step, the value $s(t_i + \Delta t)$ is refined by up to $3$ steps of Newton's method. 

For both cases of 5pt and Scranton problems, there are additional polynomial constraints which rule out certain spurious solutions.
However, one advantage of our HC method is that, although the vanishing set of the $f$ we use is strictly larger than $M,$ these \emph{additional constraints do not need to be explicitly enforced}---See SM~\cref{sec:HC-details}

\subsection{Efficient HC implementation}


\hide{HC has been applied in a number of theoretical studies that aim to classify minimal problems (eg.~\cite{Kileel-MPCTV-2016,PLMP,PL1P,duff2021galois}.)
However, its use as a solver in real-time applications is not well established.}
Our work builds on the core of an optimized HC solver introduced in~\cite{TRPLP} that was originally developed for the problem of computing the relative pose of three calibrated cameras from corresponding point-line incidences. The optimized solver in that work is globally convergent with probability $1$, but needs about $500$ miliseconds to track $312$ complex solution paths to solve a single problem instance. 

By contrast, our solver for Scranton tracks a single path in under 10 microseconds, a speedup of more than $1000 \times $.\hide{Moreover, the same method applied to five points in two views achieves roughly the same runtime as Nist\'{e}r's 5-point method, with comparable accuracy.} 
As noted in~\cref{sec:prev}, much of this dramatic speedup is because global HC methods must compute all solutions over the complex numbers, whereas our method computes one, allowing now a greater probability of failure for a given data sample.
Moreover, the start system in our HC method is tailored to the input by the anchor selection procedure.

There are also significant implementation-specific speedups.
For instance, we obtain another $\approx 14\times$ 
\hide{\anton{14 is rather large... Is this not only arithmetic, but also linear algerba?}\PH{We compare complex HC with gamma trick and real HC without gamma trick. I think the absence of gamma trick may be responsible for some of the speedup}}
speedup by performing all computations in \emph{real}, instead of complex arithmetic.
We also obtain an $\approx 5$ speedup by optimizing the linear algebra underlying predictor/corrector steps; the Jacobian matrices of our depth-formulated are sparse, leading to inexpensive closed-form solutions.\footnote{See SM~Sec.~\ref{sec:HC-details} for more details.}

\section{Minimal problem formulation}
\label{sec:formulation}
\begin{figure}
\begin{center}
\ifArXiV
\begin{tikzpicture}[scale = 1.0]
\else
\begin{tikzpicture}[scale = 0.40]
\fi

\filldraw[black] (0,0) circle[radius=2pt] node[anchor=east] {$C_i$};
\filldraw[black] (7,0) circle[radius=2pt] node[anchor=west] {$C_j$};

\filldraw[black] (3.9,3) circle[radius=2pt] node[anchor=south] {$X_m$};
\filldraw[black] (3.5,1.6) circle[radius=2pt] node[anchor=north, yshift={-0.15cm}] {$X_k$};

\draw[red, thick, ->] (0,0) -- (3.9,3);
\draw[red, thick, ->] (0,0) -- (3.5,1.6);
\draw[blue, thick, ->] (7,0) -- (3.9,3);
\draw[blue, thick, ->] (7,0) -- (3.5,1.6);

\draw[gray] (2,0) -- (0.5,2);
\draw[gray] (5,0) -- (6.5,2);

\filldraw[black!30!red] (1.5,0.65) circle[radius=1.5pt] node[anchor=north, left=0.1cm, below = 0.1cm] {$x_{k,i}$};
\filldraw[black!30!red] (1.25,0.95) circle[radius=1.5pt] node[anchor=east] {$x_{m,i}$};
\filldraw[black!30!blue] (5.5,0.7) circle[radius=1.5pt] node[anchor=north, xshift=0.2cm, yshift={-0.1cm}] {$x_{k,j}$};
\filldraw[black!30!blue] (5.85,1.1) circle[radius=1.5pt] node[anchor=west] {$x_{m,j}$};

\draw[magenta!80!blue, thick] (3.9,3) -- (3.5,1.6);

\node[black!30!red] at (1.5,2.4) {\tiny $\lambda_{m,i} v_{m,i}$};
\node[black!30!blue] at (6.1,2.4) {\tiny $\lambda_{m,j} v_{m,j}$};

\end{tikzpicture}
\ifArXiV
\begin{tikzpicture}[scale = 1.0]
\else
\begin{tikzpicture}[scale = 0.40]
\fi

\filldraw[black] (0,0.3) circle (2pt) node[anchor=east] {$C_1$};
\filldraw[black] (7,0.2) circle (2pt) node[anchor=west] {$C_3$};
\filldraw[black] (3.5,-2) circle (2pt) node[anchor=west] {$C_2$};

\filldraw[black] (3.4,2.7) circle (1.5pt) node[anchor=south] {$X_m$};
\filldraw[black] (2.2,2.3) circle (1.5pt);
\filldraw[black] (3.2,0.8) circle (1.5pt);
\filldraw[black] (5,2.6) circle (1.5pt) node[anchor=west] {$X_{1}$};
\draw[cyan, thick] (5,2) -- (5,3.2);

\draw[gray, thin] (0.5,1.1) -- (1.5,0.1);
\draw[gray, thin] (0.5,2) -- (1.5,1);
\draw[gray, thin] (0.5,0.2) -- (0.5,2);
\draw[gray, thin] (1.5,1) -- (1.5,-0.8);
\draw[gray, thin] (0.5,0.2) -- (1.5,-0.8);

\draw[gray, thin] (6.5,1.1) -- (5.5,0.1);
\draw[gray, thin] (6.5,2) -- (5.5,1);
\draw[gray, thin] (6.5,0.2) -- (6.5,2);
\draw[gray, thin] (5.5,1) -- (5.5,-0.8);
\draw[gray, thin] (5.5,-0.8) -- (6.5,0.2);

\draw[gray, thin] (2.5,-0.6) -- (4.5,-0.6);
\draw[gray, thin] (2.5,0.3) -- (4.5,0.3);
\draw[gray, thin] (2.5,-1.5) -- (2.5,0.3);
\draw[gray, thin] (4.5,-1.5) -- (4.5,0.3);
\draw[gray, thin] (4.5,-1.5) -- (2.5,-1.5);


\draw[red, very thick, ->] (0,0.3) -- (3.4, 2.7);

\draw[red!60, thick] (0,0.3) -- (5,2.6);

\draw[blue, very thick, ->] (7,0.2) -- (3.4, 2.7);

\draw[blue!60, thin] (7,0.2) -- (3.2,0.8);
\draw[blue!60, thick] (7,0.2) -- (5,2.6);

\draw[green, very thick, ->] (3.5,-2) -- (3.4, 2.7);

\draw[green!60, thin] (3.5,-2) -- (2.2,2.3);
\draw[green!60, thin] (3.5,-2) -- (3.2,0.8);
\draw[green!60, thin] (3.5,-2) -- (5,2.6);

\filldraw[black!30!red] (1,1) circle[radius=1.5pt] node[anchor=south, xshift={-0.15cm}] {\tiny $x_{m,1}$};
\filldraw[black!30!blue] (5.85,1) circle[radius=1.5pt] node[anchor=north, below=0.1cm] {\tiny $x_{m,3}$};
\filldraw[black!60!green] (3.46,-0.3) circle[radius=1.5pt] node[anchor=west, yshift=0.1cm, xshift={-0.05cm}] {\tiny $x_{m,2}$};

\filldraw[black!30!red] (1.3,0.3) circle[radius=1.5pt] node[anchor=north, xshift=-0.1cm] {\tiny $x_{1,1}$};
\draw[cyan, thick, ->] (1.3,0.3) -- (1.3,0.9);
\node[cyan] at (1.75, 0.6) {\small $l$};

\filldraw[black!30!blue] (6.36,0.97) circle[radius=1.5pt];
\filldraw[black!30!green] (3.95,-0.6) circle[radius=1.5pt];

\node[black!30!red] at (1.3,2.7) {\tiny $\lambda_{m,1} v_{m,1}$};

\end{tikzpicture}

\caption{(left) The main geometrical constraint. (right) Scranton formulation. Four points $X_m, m \in \{1,...,4\}$ are projected into three cameras $C_1$, $C_2$, $C_3$. The first point $X_1$ projects in the first camera to a point on a vertical line passing through $x_{1,1}$, with $l$ being distance from $x_{1,1}$.}
\label{fig:depth}
\end{center}
\end{figure}
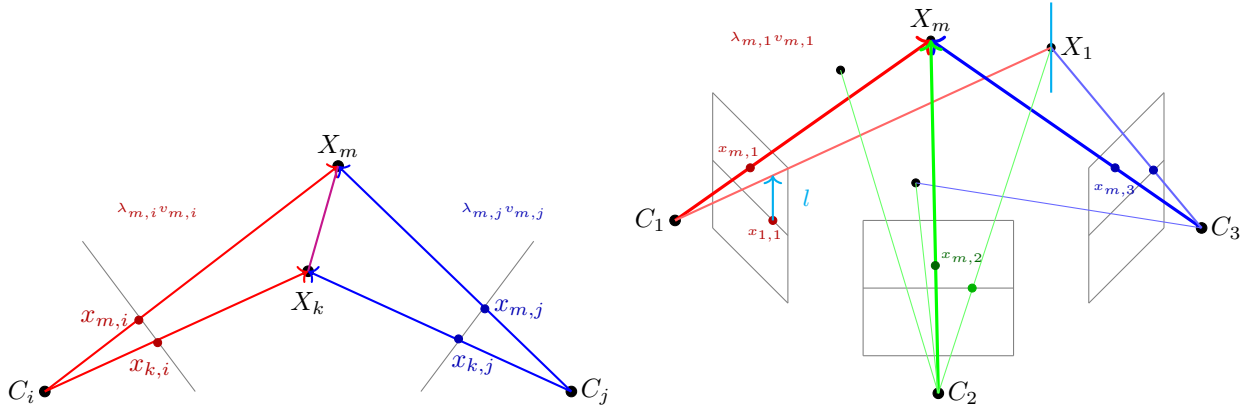

We now describe the polynomial systems used by our 5pt and 4pt HC solvers.
The unknowns in both systems are the normalized depths in each camera, as formulated in previous works~\cite{zhang2011pose, QuanTM2006}. We choose this formulation over others because (i) due to low degree and sparsity, it leads to fast evaluation of straight-line programs and fast execution of linear algebra subroutines used in homotopy tracking and (ii) it works well in tandem with our normalization procedure described in~\cref{sec:preprocess}.

To derive polynomial constraints relating depths and image points, consider two 3D points $X_k, X_m$ labeled by $k,m$ that are projected by two calibrated cameras labeled by $i,j$ into image points $x_{k,i}, x_{m,i}, x_{k,j}, x_{m,j}$ with the corresponding homogeneouos coordinates given by $v = [x;1]$. We see, Fig.~\ref{fig:depth}, that 
$
    || \lambda_{k,i} v_{k,i} - \lambda_{m,i} v_{m,i} ||^2 
  = || \lambda_{k,j} v_{k,j} - \lambda_{m,j} v_{m,j} ||^2
$
must hold, where the unknown $\lambda_{k,i}$ is the depth of the 3D point $X_k$ in camera $i$. This constraint means that the distance between every two 3D points, when reconstructed in different cameras, must be the same. 

\hide{
Let $X_i$ denote the $i$-th 3D point, $\textbf{x}_{i,j}$ the homogeneous projection of point $X_i$ into the $j$-th camera, and $\lambda_{i,j}$ the depth of point $X_i$ in the $j$-th camera. Because we assume calibrated settings, the coordinates of point $X_i$ in the coordinate system of camera $j$ can be computed as $\lambda_{i,j} \textbf{x}_{i,j}$.

Now, we are going to describe the principle of the depth formulation, in which the unknowns are the depths $\lambda_{i,j}$ and the parameters are the projections $\textbf{x}_{i,j}$. The Euclidean distance between two 3D points $X_i, X_{i'}$ can be computed from the projections and depths of the points in any camera $j$ as:
\begin{equation}
    || X_i - X_{i'} || = || \lambda_{i,j} \textbf{x}_{i,j} - \lambda_{i',j} \textbf{x}_{i',j} ||
\end{equation}
Let us have two different cameras $j$, $j'$. The Euclidean distance computed from both cameras has to be equal. The situation is depicted in Figure \ref{fig:depth}. This relation produces one polynomial equation with the depths as unknowns:
\begin{equation}
    || \lambda_{i,j} \textbf{x}_{i,j} - \lambda_{i',j} \textbf{x}_{i',j} ||^2 = || \lambda_{i,j'} \textbf{x}_{i,j'} - \lambda_{i',j'} \textbf{x}_{i',j'} ||^2\label{eq:depth_base}
\end{equation}
If we take equations for every pair of points $i$, $i'$, and for every pair of cameras $j$, $j'$, we obtain a system of polynomial equations with a finite nonzero number of solutions. This system can be reduced to a square system and solved by HC. Then, the relative pose can be recovered.}

\subsection{5pt minimal problem}\label{sec:5pt}
The 5pt problem is parametrized by the projection of 5 3D points into 2 calibrated cameras. There are $10$ unknown depths $\lambda_{i,j}$, $i=1,\ldots,5$, $j=1,2$, and $10 = \binom{5}{2}$ equations
\begin{equation}
    || \lambda_{k,1} v_{k,1} - \lambda_{m,1} v_{m,1} ||^2 
  = || \lambda_{k,2} v_{k,2} - \lambda_{m,2} v_{m,2} ||^2
  \label{eq:5pt-equations}
\end{equation}
$k,m=1,\ldots,5$, $k\neq m$. To dehomogenize this system, we set $\lambda_{1,1}=1$ to obtain a system of $10$ equations in $9$ unknowns. It has $80$ solutions for generic parameters $v_{i,j}$. There are two isolated  singular solutions $\lambda = [1,0,0,0,0;\pm a,0,0,0,0]$, with multiplicity $20$, and $40$ isolated nonsingular solutions\hide{with sign symmetry, i.e., for a solution $\lambda$, there is also the solutions $-\lambda$}. Among the $40$ nonsingular solutions, there are at most $10$ with all depths positive which extend to a rotation with $\det R_2 = 1$.

For our HC solver, we have to select a square subsystem, i.e., 9 equations for 9 unknowns. 
For generic parameters, any equation can be dropped to get a square system with $160$ solutions, where the two singular solutions have multiplicity $32$ and the number of nonsingular solutions rises to $96$.  
As noted in~\cref{sec:hc_solving}, the dropped equation and other polynomial constraints ($\det R_2 = 1$) need not be explicitly enforced by our HC solver.\footnote{See SM Sec.~\ref{sec:5pt-formulation-details} for additional details about the problem formulation.}.

\hide{
We fix the first depth $\lambda_{1,1}=1$ to prevent the scale ambiguity. Depth formulation of the 5 point problem is a system of $9$ polynomial equations in $9$ unknowns:

\begin{equation}
\begin{split}
    \left\lVert \lambda_{i,1} \textbf{x}_{i,1} - \lambda_{i',1} \textbf{x}_{i',1} \right\rVert^2 = \left\lVert \lambda_{i,2} \textbf{x}_{i,2} - \lambda_{i',2} \textbf{x}_{i',2} \right\rVert^2,
    \\ (i,i') \in \{(1,2), (1,3), (1,4), (1,5), (2,3), (2,4), (2,5), (3,4), (3,5)\} \label{eq:depth_square}
\end{split}
\end{equation}
}

\subsection{Scranton relaxation of the 4pt problem}
\label{sec:4pt4vFormulation}

The 4pt problem consists of the projection of 4 points into 3 calibrated cameras. 
Therefore, it involves $12$ unknown depths $\lambda_{i,j}$, $i=1,2,3,4$, $j=1,2,3$, and $18~=~\binom{4}{2} \binom{3}{2}$ equations.  However, only 2 equations, from each $\binom{3}{2}$ equations involving the same pair of 3D points, are independent. Hence we get $12$ equations
\begin{equation}
    || \lambda_{k,i} v_{k,i} - \lambda_{m,i} v_{m,i} ||^2 
  = || \lambda_{k,j} v_{k,j} - \lambda_{m,j} v_{m,j} ||^2
  \label{eq:4pt-equations}
\end{equation}
$k,m=1,\ldots,4$, $k\neq m$ and, e.g., $i=1,2$, $j=i+1$. To dehomogenize the system, we set $\lambda_{1,1}=1$. Unlike for the 5pt problem, here we get an overconstrained system of $12$ equations for $11$ unknown depths, which has no solution for generic (noisy) parameters $v_{k,i}$.  To get a minimal problem, we replace $v_{1,1}$ by $v_{1,1} + l\,[0;1;0]$, where $l$ is a new unknown. 
This relaxation allows us to ``adjust'' the second coordinate of the first point in the first camera. 
Notice that by relaxing the first point in the first view, which has $\lambda_{1,1} = 1$, the equations involving that point
\begin{equation*}
    || v_{1,1} + l\,[0;1;0] - \lambda_{m,1} v_{m,1} ||^2 
  = || \lambda_{1,2} v_{1,2} - \lambda_{m,2} v_{m,2} ||^2
\end{equation*}
for $m=2,3,4$ remain quadratic in unknowns $(\lambda , l).$
Solutions to the minimal problem ``Scranton", Fig.~\ref{fig:depth}, are solutions to this square, inhomogeneous system of $12$ polynomials used in our HC solver.\footnote{See SM Sec.~\ref{sec:Scranton-formulation-details} for additional details about Scranton.}.

\section{Problem preprocessing}
\label{sec:preprocess}
To simplify both the learning the anchor selection strategy $\sigma $ and the HC tracking, we considered several schemes for \emph{normalizing} the input image correspondences, i.e., the parameters $p$ of the problems.
Our chosen normalization yields single representative $p$ for all problems that differ from $p$ up to camera re-orientation or permutation of cameras or correspondences. Once the normalized problem is solved, the original problem may be solved by applying a transformation $R_i^{-1}$ described below. \hide{The normalization makes the problems more similar to each other, and therefore, tracking between two normalized problems is faster and more successful than tracking between original problems. Moreover, the normalized problems live in a lower-dimensional subspace, and thus the selection of anchors is a simpler task. The normalization has to be fast because it is done in the online phase for every input problem $p$.}

For the 5pt problem, a problem $p$ is given by image coordinates $x_{i,j} \in \RR^2$ with cameras indexed by $i=1,2$ and points indexed by $j=1,\ldots,5$. First, we construct unit 3D vectors representing the rays of the image points as $v_{i,j} = [x_{i,j};1]/||[x_{i,j};1]||$. Next, we compute the mean ray for each camera $m_i = \mbox{mean}_j(v_{i,j})$. Then, we find the ray $v_{i^*j^*}$ that contains the largest angle with the mean ray $m_i$ of its camera, i.e., $(i^*,j^*) = \mbox{argmax}_{(i,j)} \angle(x_{i,j},m_i)$. Next, we compute $w_{i,j} = R_i v_{i,j}$ such that $R_i m_i = [0;0;1]$ and $y_{i^*,j^*}$, as well as the corresponding $y_{1-i^*,j^*}$, have the second coordinate equal to $0$, i.e., we put them on the ``$x$ axis''. Finally, we swap the cameras to make the camera $i^*$ the first one, project 3D rays $w_{i,j}$ back to the image points $x_{i,j}=w_{i,j}/w_{i,j}^{(3)}$, and reorder the image correspondences counterclockwise starting with $j^*$\footnote{See SM Fig.~\ref{fig:normalized-5pt} for an example of image correspondences after the normalization, and SM Sec.~\ref{supp:normalizations} for a comparison with other normalizations which delivered worse results in our evaluation.}. 

For the 4pt problem, cameras are ordered according to angles of the first point; $\angle([x_{1,1};1],[0,0,1]) \geq \angle([x_{2,1};1],[0,0,1]) \geq \angle([x_{3,1};1],[0,0,1]).$

\section{Experiments - RANSAC evaluation}
\label{sec:experiments}
\hide{\tim{Right now, ``Experiments" section only reports on 5pt solver.
Should we move Table 4 here?
}}

\begin{figure}[t]
    \centering
    \includegraphics[width=0.7\linewidth]{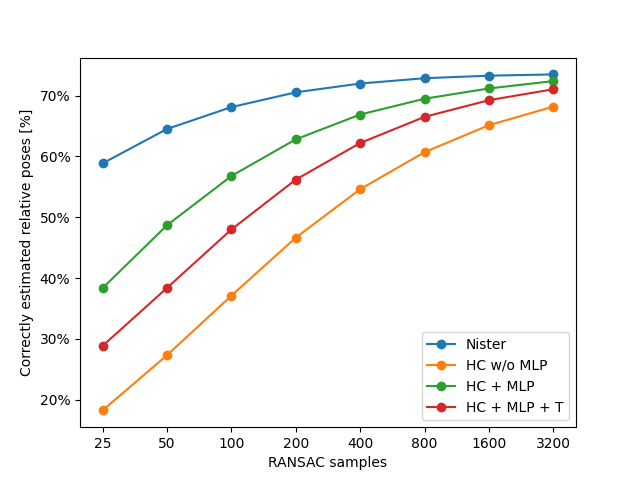}
    \caption{Percentage of camera pairs from the 2020 RANSAC Tutorial \cite{ransac_tutorial} for which the relative pose obtained by RANSAC has rotation and translation error less than $10^\circ$.}
    \label{fig:rot_err}
\end{figure}

\hide{\subsection{Evaluation in RANSAC scheme}}
\hide{Consider HC solver for the 5pt problem with anchors $A$ (Sec.~\ref{sec:generate_anchors}) and an MLP classifier (Sec.~\ref{sec:start-p-s}).} To show how our method generalizes to different real scenes and to data contaminated by noise and wrong matches,
we evaluate our approach for the 5pt problem on the dataset from the CVPR 2020 RANSAC Tutorial~\cite{ransac_tutorial} consising of 2 validation scenes and 11 test scenes, each comprising 4950 camera pairs. For every camera pair, a set of matched 2D points is known. The points are contaminated with noise and mismatches. We evaluate solvers by plugging them into a RANSAC scheme~\cite{Raguram-USAC-PAMI-2013} and computing relative poses for camera pairs in each scene. We evaluate the rotation error and translation error separately. Our evaluation metric is the percentage of relative poses whose angular distance from the ground truth is less than $10^\circ$. 
\arxiv{We believe that this metric is justified, because the main purpose of the RANSAC procedure is to separate the correct matches from the mismatches, and a more precise relative pose can be obtained by local optimization on the inliers.}


We consider our HC solver with a single anchor, our HC solver with MLP without trash, and our HC solver with MLP and trash. To estimate the success rate of our solver on this data, we compare it with the Nistér 5 point solver \cite{Nister-5pt-PAMI-2004}. The success rate of the Nistér solver is close to $100 \%$, and its errors are only due to the noise and mismatches in the data. Therefore, if the success rate of a solver on the given data is, e.g., $25 \%$, we expect it to need 4 times the number of samples used for the Nistér solver to get the same results. We have considered RANSAC with $25, 50, 100, 200, 400, 800, 1600,$ and $3200$ samples. The inlier ratio is $3$px. The relation between the number of samples and the percentage of correctly estimated cameras is shown in Fig.~\ref{fig:rot_err}. The graph shows that the lower success rate of our method can be compensated by running RANSAC for more samples. Our method with MLP requires about 4 times more samples than the Nistér solver. Therefore, the success rate on the data from \cite{ransac_tutorial} is around $25 \%$, which is about $1.6$ times lower than the success rate on the testing data from ETH 3D dataset 
\footnote{See SM~Sec.~\ref{sec:ablations} for the study of our engineering choices.}.\\

\section{Conclusion}
Our approach to  solving  hard  minimal problems for RANSAC  framework, \arxiv{which uses efficient homotopy continuation and machine learning to avoid solving for many spurious solutions,} is fast and delivers correct results. Supplementary Material (SM) presents more details and experiments. Our code and and data \ifArXiV are available at https://github.com/petrhruby97/learning\_minimal \fi

{\noindent \bf Limitations of our approach:} 
First, we sacrifice the high success rate of a complex HC method for a fast, real HC method that fails more frequently. Nevertheless, when combined with trained models, our method succeeds in computing real solutions often enough to be useful in RANSAC. Secondly, our MLP model represents only what it is trained for. Still, we saw that it was able to represent real data distributions while keeping small size and fast evaluation. Fitting to a particular data distribution may also be useful in special situations, e.g., when cameras are mounted on a vehicle, hence having special motions. 

{\small
\bibliographystyle{ieee_fullname}
\bibliography{local}

\begin{thebibliography}{10}\itemsep=-1pt

\bibitem{AgarwalLST17}
Sameer Agarwal, Hon{-}leung Lee, Bernd Sturmfels, and Rekha~R. Thomas.
\newblock On the existence of epipolar matrices.
\newblock {\em International Journal of Computer Vision}, 121(3):403--415,
  2017.

\bibitem{AholtO14}
Chris Aholt and Luke Oeding.
\newblock The ideal of the trifocal variety.
\newblock {\em Math. Comput.}, 83(289):2553--2574, 2014.

\bibitem{Alismail-odometry}
Hatem~Said Alismail, Brett Browning, and M~Bernardine Dias.
\newblock Evaluating pose estimation methods for stereo visual odometry on
  robots.
\newblock In {\em the 11th International Conference on Intelligent Autonomous
  Systems (IAS-11)}, January 2011.

\bibitem{Barath-CVPR-2018}
Daniel Barath.
\newblock Five-point fundamental matrix estimation for uncalibrated cameras.
\newblock In {\em 2018 {IEEE} Conference on Computer Vision and Pattern
  Recognition, {CVPR} 2018, Salt Lake City, UT, USA, June 18-22, 2018}, pages
  235--243, 2018.

\bibitem{Barath-TIP-2018}
Daniel Barath and Levente Hajder.
\newblock Efficient recovery of essential matrix from two affine
  correspondences.
\newblock {\em {IEEE} Trans. Image Processing}, 27(11):5328--5337, 2018.

\bibitem{Barath-CVPR-2017}
Daniel Barath, Tekla Toth, and Levente Hajder.
\newblock A minimal solution for two-view focal-length estimation using two
  affine correspondences.
\newblock In {\em 2017 {IEEE} Conference on Computer Vision and Pattern
  Recognition, {CVPR} 2017, Honolulu, HI, USA, July 21-26, 2017}, pages
  2557--2565, 2017.

\bibitem{DBLP:books/daglib/0032895}
Daniel~J. Bates, Andrew~J. Sommese, Jonathan~D. Hauenstein, and Charles~W.
  Wampler.
\newblock {\em Numerically Solving Polynomial Systems with Bertini}, volume~25
  of {\em Software, environments, tools}.
\newblock {SIAM}, 2013.

\bibitem{DBLP:conf/cvpr/BhayaniKH20}
Snehal Bhayani, Zuzana Kukelova, and Janne Heikkil{\"{a}}.
\newblock A sparse resultant based method for efficient minimal solvers.
\newblock In {\em 2020 {IEEE/CVF} Conference on Computer Vision and Pattern
  Recognition, {CVPR} 2020, Seattle, WA, USA, June 13-19, 2020}, pages
  1767--1776. {IEEE}, 2020.

\bibitem{opencv_library}
G. Bradski.
\newblock {The OpenCV Library}.
\newblock {\em Dr. Dobb's Journal of Software Tools}, 2000.

\bibitem{DBLP:conf/icms/BreidingT18}
Paul Breiding and Sascha Timme.
\newblock Homotopycontinuation.jl: {A} package for homotopy continuation in
  julia.
\newblock In James~H. Davenport, Manuel Kauers, George Labahn, and Josef Urban,
  editors, {\em Mathematical Software - {ICMS} 2018 - 6th International
  Conference, South Bend, IN, USA, July 24-27, 2018, Proceedings}, volume 10931
  of {\em Lecture Notes in Computer Science}, pages 458--465. Springer, 2018.

\bibitem{Byrod-ECCV-2008}
Martin Byr{\"o}d, Klas Josephson, and Kalle {\AA}str{\"o}m.
\newblock A column-pivoting based strategy for monomial ordering in numerical
  {G}r\"{o}bner basis calculations.
\newblock In {\em European Conference on Computer Vision (ECCV)}, volume 5305,
  pages 130--143. Springer, 2008.

\bibitem{DBLP:conf/eccv/CamposecoSP16}
F. Camposeco, T. Sattler, and M. Pollefeys.
\newblock Minimal solvers for generalized pose and scale estimation from two
  rays and one point.
\newblock In {\em {ECCV} -- European Conference on Computer Vision}, pages
  202--218, 2016.

\bibitem{Cox-UAG-1998}
David Cox, John Little, and Donald O'Shea.
\newblock {\em Using Algebraic Geometry}.
\newblock Springer, 1998.

\bibitem{Duff-Monodromy}
Timothy Duff, Cvetelina Hill, Anders Jensen, Kisun Lee, Anton Leykin, and Jeff
  Sommars.
\newblock Solving polynomial systems via homotopy continuation and monodromy.
\newblock {\em IMA Journal of Numerical Analysis}, 2018.

\bibitem{PLMP}
T. {Duff}, K. {Kohn}, A. {Leykin}, and T. {Pajdla}.
\newblock {PLMP} - point-line minimal problems in complete multi-view
  visibility.
\newblock In {\em 2019 IEEE/CVF International Conference on Computer Vision
  (ICCV)}, pages 1675--1684, 2019.

\bibitem{PL1P}
Timothy Duff, Kathl{\'{e}}n Kohn, Anton Leykin, and Tom{\'{a}}s Pajdla.
\newblock P{L\({}_{\mbox{1}}\)P} - point-line minimal problems under partial
  visibility in three views.
\newblock In Andrea Vedaldi, Horst Bischof, Thomas Brox, and Jan{-}Michael
  Frahm, editors, {\em Computer Vision - {ECCV} 2020 - 16th European
  Conference, Glasgow, UK, August 23-28, 2020, Proceedings, Part {XXVI}},
  volume 12371 of {\em Lecture Notes in Computer Science}, pages 175--192.
  Springer, 2020.

\bibitem{duff2021galois}
Timothy Duff, Viktor Korotynskiy, Tomas Pajdla, and Margaret~H Regan.
\newblock Galois/monodromy groups for decomposing minimal problems in 3d
  reconstruction.
\newblock {\em arXiv preprint arXiv:2105.04460}, 2021.

\bibitem{Elqursh-CVPR-2011}
Ali Elqursh and Ahmed~M. Elgammal.
\newblock Line-based relative pose estimation.
\newblock In {\em The 24th {IEEE} Conference on Computer Vision and Pattern
  Recognition, {CVPR} 2011, Colorado Springs, CO, USA, 20-25 June 2011}, pages
  3049--3056. {IEEE} Computer Society, 2011.

\bibitem{DBLP:journals/corr/abs-1201-5810}
Ioannis~Z. Emiris.
\newblock A general solver based on sparse resultants.
\newblock {\em CoRR}, abs/1201.5810, 2012.

\bibitem{TRPLP}
Ricardo Fabbri, Timothy Duff, Hongyi Fan, Margaret~H. Regan, David da~Costa~de
  Pinho, Elias~P. Tsigaridas, Charles~W. Wampler, Jonathan~D. Hauenstein,
  Peter~J. Giblin, Benjamin~B. Kimia, Anton Leykin, and Tom{\'{a}}s Pajdla.
\newblock {TRPLP} - {T}rifocal relative pose from lines at points.
\newblock In {\em 2020 {IEEE/CVF} Conference on Computer Vision and Pattern
  Recognition, {CVPR} 2020, Seattle, WA, USA, June 13-19, 2020}, pages
  12070--12080. {IEEE}, 2020.

\bibitem{Fabbri-PAMI-2020}
R. {Fabbri}, P. {Giblin}, and B. {Kimia}.
\newblock Camera pose estimation using first-order curve differential geometry.
\newblock {\em IEEE Transactions on Pattern Analysis and Machine Intelligence},
  2020.

\bibitem{DBLP:conf/issac/FaugereMRD08}
Jean{-}Charles Faug{\`{e}}re, Guillaume Moroz, Fabrice Rouillier, and Mohab
  Safey~El Din.
\newblock Classification of the perspective-three-point problem, discriminant
  variety and real solving polynomial systems of inequalities.
\newblock In J.~Rafael Sendra and Laureano Gonz{\'{a}}lez{-}Vega, editors, {\em
  Symbolic and Algebraic Computation, International Symposium, {ISSAC} 2008,
  Linz/Hagenberg, Austria, July 20-23, 2008, Proceedings}, pages 79--86. {ACM},
  2008.

\bibitem{Fischler-Bolles-ACM-1981}
M.~A. Fischler and R.~C. Bolles.
\newblock Random sample consensus: a paradigm for model fitting with
  applications to image analysis and automated cartography.
\newblock {\em Commun. ACM}, 24(6):381--395, 1981.

\bibitem{GriffHarr}
Phillip Griffiths and Joseph Harris.
\newblock {\em Principles of algebraic geometry}.
\newblock Wiley Classics Library. John Wiley \& Sons, Inc., New York, 1994.
\newblock Reprint of the 1978 original.

\bibitem{Grunert-1841}
J.~A. Grunert.
\newblock Das pothenotische {P}roblem in erweiterter {G}estalt nebst \"uber
  seine {A}nwendungen in {G}eod\"asie.
\newblock {\em In Grunerts Archiv f\"ur Mathematik und Physik}, 1841.

\bibitem{DBLP:journals/ijcv/HaralickLON94}
Robert~M. Haralick, Chung{-}Nan Lee, Karsten Ottenberg, and Michael
  N{\"{o}}lle.
\newblock Review and analysis of solutions of the three point perspective pose
  estimation problem.
\newblock {\em Int. J. Comput. Vis.}, 13(3):331--356, 1994.

\bibitem{Hartley-PAMI-2012}
R. Hartley and Hongdong Li.
\newblock An efficient hidden variable approach to minimal-case camera motion
  estimation.
\newblock {\em IEEE PAMI}, 34(12):2303--2314, 2012.

\bibitem{HZ-2003}
Richard Hartley and Andrew Zisserman.
\newblock {\em Multiple View Geometry in Computer Vision}.
\newblock Cambridge, 2nd edition, 2003.

\bibitem{DBLP:conf/iccv/Heikkila17}
Janne Heikkil{\"{a}}.
\newblock Using sparse elimination for solving minimal problems in computer
  vision.
\newblock In {\em {IEEE} International Conference on Computer Vision, {ICCV}
  2017, Venice, Italy, October 22-29, 2017}, pages 76--84. {IEEE} Computer
  Society, 2017.

\bibitem{Holt-PAMI-1995}
Robert~J. Holt and Arun~N. Netravali.
\newblock Uniqueness of solutions to three perspective views of four points.
\newblock {\em {IEEE} Trans. Pattern Anal. Mach. Intell.}, 17(3):303--307,
  1995.

\bibitem{Kileel-MPCTV-2016}
Joe Kileel.
\newblock Minimal problems for the calibrated trifocal variety.
\newblock {\em SIAM Journal on Applied Algebra and Geometry}, 1(1):575--598,
  2017.

\bibitem{DBLP:conf/cvpr/KneipSS11}
L. Kneip, D. Scaramuzza, and R. Siegwart.
\newblock A novel parametrization of the perspective-three-point problem for a
  direct computation of absolute camera position and orientation.
\newblock In {\em {CVPR} -- {IEEE} Conference on Computer Vision and Pattern
  Recognition}, pages 2969--2976, 2011.

\bibitem{DBLP:conf/eccv/KneipSP12}
L. Kneip, R. Siegwart, and M. Pollefeys.
\newblock Finding the exact rotation between two images independently of the
  translation.
\newblock In {\em {ECCV} -- European Conference on Computer Vision}, pages
  696--709, 2012.

\bibitem{Kuang-ICCV-2013}
Yubin Kuang and Kalle {\AA}str{\"{o}}m.
\newblock Pose estimation with unknown focal length using points, directions
  and lines.
\newblock In {\em {IEEE} International Conference on Computer Vision, {ICCV}
  2013, Sydney, Australia, December 1-8, 2013}, pages 529--536, 2013.

\bibitem{kuang-astrom-2espc2-13}
Yubin Kuang and Kalle {\AA}str{\"o}m.
\newblock Stratified sensor network self-calibration from {TDOA} measurements.
\newblock In {\em 21st European Signal Processing Conference}, 2013.

\bibitem{kukelova2008automatic}
Zuzana Kukelova, Martin Bujnak, and Tomas Pajdla.
\newblock Automatic generator of minimal problem solvers.
\newblock In {\em European Conference on Computer Vision (ECCV)}, 2008.

\bibitem{Kukelova-PolyEig-PAMI-2012}
Zuzana Kukelova, Martin Bujnak, and Tomas Pajdla.
\newblock Polynomial eigenvalue solutions to minimal problems in computer
  vision.
\newblock {\em IEEE Transactions on Pattern Analysis and Machine Intelligence},
  2012.

\bibitem{PoseLib}
Viktor Larsson.
\newblock {PoseLib - Minimal Solvers for Camera Pose Estimation}, 2020.

\bibitem{larsson2017efficient}
Viktor Larsson, Kalle {\AA}str{\"o}m, and Magnus Oskarsson.
\newblock Efficient solvers for minimal problems by syzygy-based reduction.
\newblock In {\em Computer Vision and Pattern Recognition (CVPR)}, 2017.

\bibitem{Larsson-Saturated-ICCV-2017}
Viktor Larsson, Kalle {\AA}str{\"{o}}m, and Magnus Oskarsson.
\newblock Polynomial solvers for saturated ideals.
\newblock In {\em {IEEE} International Conference on Computer Vision, {ICCV}
  2017, Venice, Italy, October 22-29, 2017}, pages 2307--2316, 2017.

\bibitem{larsson2017making}
Viktor Larsson, Zuzana Kukelova, and Yinqiang Zheng.
\newblock Making minimal solvers for absolute pose estimation compact and
  robust.
\newblock In {\em International Conference on Computer Vision (ICCV)}, 2017.

\bibitem{Larsson-CVPR-2018}
Viktor Larsson, Magnus Oskarsson, Kalle {\AA}str{\"{o}}m, Alge Wallis, Zuzana
  Kukelova, and Tom{\'{a}}s Pajdla.
\newblock Beyond grobner bases: Basis selection for minimal solvers.
\newblock In {\em 2018 {IEEE} Conference on Computer Vision and Pattern
  Recognition, {CVPR} 2018, Salt Lake City, UT, USA, June 18-22, 2018}, pages
  3945--3954, 2018.

\bibitem{leykin2011numerical}
Anton Leykin.
\newblock Numerical algebraic geometry.
\newblock {\em Journal of Software for Algebra and Geometry}, 3(1):5--10, 2011.

\bibitem{Miraldo-ECCV-2018}
Pedro Miraldo, Tiago Dias, and Srikumar Ramalingam.
\newblock A minimal closed-form solution for multi-perspective pose estimation
  using points and lines.
\newblock In {\em Computer Vision - {ECCV} 2018 - 15th European Conference,
  Munich, Germany, September 8-14, 2018, Proceedings, Part {XVI}}, pages
  490--507, 2018.

\bibitem{mirzaei2011optimal}
Faraz~M Mirzaei and Stergios~I Roumeliotis.
\newblock Optimal estimation of vanishing points in a manhattan world.
\newblock In {\em International Conference on Computer Vision (ICCV)}, 2011.

\bibitem{morgan2009solving}
A. Morgan.
\newblock {\em Solving Polynomial Systems Using Continuation for Engineering
  and Scientific Problems}.
\newblock Classics in Applied Mathematics. Society for Industrial and Applied
  Mathematics (SIAM, 3600 Market Street, Floor 6, Philadelphia, PA 19104),
  2009.

\bibitem{ransac_tutorial}
Dmytro Myshkin.
\newblock Benchmarking robust estimation methods.
\newblock {\em Tutorial “RANSAC in 2020”, CVPR}, 2020.

\bibitem{Nister-5pt-PAMI-2004}
D. Nist\'er.
\newblock An efficient solution to the five-point relative pose problem.
\newblock {\em IEEE Transactions on Pattern Analysis and Machine Intelligence},
  26(6):756--770, June 2004.

\bibitem{Nister04visualodometry}
David Nist{\'{e}}r, Oleg Naroditsky, and James Bergen.
\newblock Visual odometry.
\newblock In {\em Computer Vision and Pattern Recognition (CVPR)}, pages
  652--659, 2004.

\bibitem{Nister-IJCV-2006}
David Nist{\'{e}}r and Frederik Schaffalitzky.
\newblock Four points in two or three calibrated views: Theory and practice.
\newblock {\em International Journal of Computer Vision}, 67(2):211--231, 2006.

\bibitem{DBLP:conf/eccv/PerssonN18}
Mikael Persson and Klas Nordberg.
\newblock Lambda twist: An accurate fast robust perspective three point {(P3P)}
  solver.
\newblock In Vittorio Ferrari, Martial Hebert, Cristian Sminchisescu, and Yair
  Weiss, editors, {\em Computer Vision - {ECCV} 2018 - 15th European
  Conference, Munich, Germany, September 8-14, 2018, Proceedings, Part {IV}},
  volume 11208 of {\em Lecture Notes in Computer Science}, pages 334--349.
  Springer, 2018.

\bibitem{QuanTM2006}
Long Quan, Bill Triggs, and Bernard Mourrain.
\newblock Some results on minimal euclidean reconstruction from four points.
\newblock {\em Journal of Mathematical Imaging and Vision}, 24(3):341--348,
  2006.

\bibitem{Raguram-USAC-PAMI-2013}
R. Raguram, O. Chum, M. Pollefeys, J. Matas, and J.{-}M. Frahm.
\newblock {USAC:} {A} universal framework for random sample consensus.
\newblock {\em {IEEE} Transactions on Pattern Analysis Machine Intelligence},
  35(8):2022--2038, 2013.

\bibitem{DBLP:conf/cvpr/RamalingamS08}
S. Ramalingam and P.~F. Sturm.
\newblock Minimal solutions for generic imaging models.
\newblock In {\em {CVPR} -- {IEEE} Conference on Computer Vision and Pattern
  Recognition}, 2008.

\bibitem{rocco2018neighbourhood}
Ignacio Rocco, Mircea Cimpoi, Relja Arandjelović, Akihiko Torii, Tomas Pajdla,
  and Josef Sivic.
\newblock Neighbourhood consensus networks, 2018.

\bibitem{SalaunMM-ECCV-2016}
Yohann Sala{\"{u}}n, Renaud Marlet, and Pascal Monasse.
\newblock Robust and accurate line- and/or point-based pose estimation without
  manhattan assumptions.
\newblock In {\em European Conference on Computer Vision (ECCV)}, 2016.

\bibitem{Sattler-PAMI-2017}
Torsten Sattler, Bastian Leibe, and Leif Kobbelt.
\newblock Efficient {\&} effective prioritized matching for large-scale
  image-based localization.
\newblock {\em {IEEE} Trans. Pattern Anal. Mach. Intell.}, 39(9):1744--1756,
  2017.

\bibitem{saurer2015minimal}
Olivier Saurer, Marc Pollefeys, and Gim~Hee Lee.
\newblock A minimal solution to the rolling shutter pose estimation problem.
\newblock In {\em Intelligent Robots and Systems (IROS), 2015 IEEE/RSJ
  International Conference on}, pages 1328--1334. IEEE, 2015.

\bibitem{schoenberger2016sfm}
Johannes~Lutz Sch\"{o}nberger and Jan-Michael Frahm.
\newblock Structure-from-motion revisited.
\newblock In {\em Conference on Computer Vision and Pattern Recognition
  (CVPR)}, 2016.

\bibitem{schoeps2017cvpr}
Thomas Sch\"ops, Johannes~L. Sch\"onberger, Silvano Galliani, Torsten Sattler,
  Konrad Schindler, Marc Pollefeys, and Andreas Geiger.
\newblock A multi-view stereo benchmark with high-resolution images and
  multi-camera videos.
\newblock In {\em Conference on Computer Vision and Pattern Recognition
  (CVPR)}, 2017.

\bibitem{Snavely-SIGGRAPH-2006}
N. Snavely, S.~M. Seitz, and R. Szeliski.
\newblock {Photo tourism: exploring photo collections in 3D}.
\newblock In {\em ACM SIGGRAPH}, 2006.

\bibitem{snavely2008modeling}
Noah Snavely, Steven~M Seitz, and Richard Szeliski.
\newblock Modeling the world from internet photo collections.
\newblock {\em International Journal of Computer Vision (IJCV)},
  80(2):189--210, 2008.

\bibitem{DBLP:books/daglib/0014410}
Andrew~J. Sommese and Charles W.~Wampler II.
\newblock {\em The numerical solution of systems of polynomials - arising in
  engineering and science}.
\newblock World Scientific, 2005.

\bibitem{Stewenius-ISPRS-2006}
H. Stewenius, C. Engels, and D. Nist\'er.
\newblock Recent developments on direct relative orientation.
\newblock {\em ISPRS J. of Photogrammetry and Remote Sensing}, 60:284--294,
  2006.

\bibitem{Sturmfels-CBMS-2002}
Bernd Sturmfels.
\newblock {\em Solving Systems of Polynomial Equations}, volume~97 of {\em CBMS
  Regional Conferences Series}.
\newblock Amer.Math.Soc., Providence, Rhode Island, 2002.

\bibitem{taira2018inloc}
Hajime Taira, Masatoshi Okutomi, Torsten Sattler, Mircea Cimpoi, Marc
  Pollefeys, Josef Sivic, Tomas Pajdla, and Akihiko Torii.
\newblock {InLoc}: Indoor visual localization with dense matching and view
  synthesis.
\newblock In {\em {CVPR}}, 2018.

\bibitem{ventura2015efficient}
Jonathan Ventura, Clemens Arth, and Vincent Lepetit.
\newblock An efficient minimal solution for multi-camera motion.
\newblock In {\em International Conference on Computer Vision (ICCV)}, pages
  747--755, 2015.

\bibitem{DBLP:journals/cca/Verschelde10}
Jan Verschelde.
\newblock Polynomial homotopy continuation with phcpack.
\newblock {\em {ACM} Commun. Comput. Algebra}, 44(3/4):217--220, 2010.

\bibitem{zhang2011pose}
Ji Zhang, Mireille Boutin, and Daniel~G Aliaga.
\newblock Pose-free structure from motion using depth from motion constraints.
\newblock {\em IEEE transactions on image processing}, 20(10):2937--2953, 2011.

\end{thebibliography}
}
\clearpage\newpage
\renewcommand{\textfraction}{0.05}
\renewcommand{\topfraction}{0.8}
\renewcommand{\bottomfraction}{0.8}
\renewcommand{\textfraction}{0.1}
\renewcommand{\floatpagefraction}{0.8}

\ifArXiV
\begin{center}
{\Large Supplementary Material}
\end{center}
\fi

\noindent Here we give additional details, including an analysis of the 5pt and Scranton minimal problem solutions, details of our efficient homotopy continuation implementation, and experiments justifying our engineering choices.
\ifArXiV
Our code is available at https://github.com/petrhruby97/learning\_minimal
\fi

\section{A classical example of a minimal problem}
\label{supp:mpcvEx}
A classical, easy, but still essential, minimal problem in computer vision is computing the pose of a calibrated perspective camera~\cite{HZ-2003} from three points in space and their image projections~\cite{Grunert-1841,DBLP:journals/ijcv/HaralickLON94, DBLP:conf/cvpr/KneipSS11,larsson2017making,DBLP:conf/eccv/PerssonN18}. In one of its classical formulations~\cite{Grunert-1841}, it leads to a polynomial system of three equations
\begin{eqnarray*}
||X_1-X_2||^2=||\lambda_1 u_1-\lambda_2 u_2||^2 \\
||X_2-X_3||^2=||\lambda_2 u_2-\lambda_3 u_3||^2 \\
||X_3-X_1||^2=||\lambda_3 u_3-\lambda_1 u_1||^2
\end{eqnarray*}
of degree two in three unknown depths $\lambda_1, \lambda_2,\lambda_3$. Parameters of the problem are three 3D points $X_i \in \RR^3$ and homogeneous coordinates $u_i \in \RR^2$ of three image projections, altogether on $3 \times 3 + 3 \times 2 = 15$ parameters. For generic data, the system has eight complex solutions for $\lambda$'s with up to eight real solutions~\cite{DBLP:conf/issac/FaugereMRD08}. However, often, there are only zero, two, or four real solutions with positive $\lambda$'s. 

This example illustrates a typical situation occurring in minimal problem solving. The minimal problem obtained by relaxing a geometrical optimization problem, which has one optimal solution, brings in seven additional (spurious) solutions. In this case, there are always at least two real solutions corresponding to seeing the three points from the opposite sides of the plain they span.

\section{Interesting hard minimal problems}
\label{supp:important-mps} 
Recent results~\cite{PLMP, PL1P} suggest that solving minimal problems with many complex solutions is interesting. A complete classification of minimal problems for points, lines, and their incidences in the calibrated multi-view geometry appeared for the case of complete multi-view visibility~\cite{PLMP}. It has been found that there are only 30 minimal problems in that setting, but it also became clear that problems involving more than two cameras are hard for the current symbolic-numeric and homotopy continuation solvers. The number of solutions for three views starts with 64, interesting cases have 200+ solutions, and 5-view cases have 10000+ solutions. Allowing for occlusion or missed detection in images leads to even harder problems. The follow-up work~\cite{PL1P} developed a complete classification of minimal problems for generic arrangements of points and lines in space, observed partially by three calibrated perspective cameras when each line is incident to at most one point. It has been found that there is an infinite number of such minimal problems arranged into 74575 equivalence classes when caring only about camera configurations. Interestingly, this classification involves all calibrated trifocal geometry of computer vision for nonincident points and lines in space. Out of 74575 classes, only 759 classes have less than 300 solutions. The rest have (many) more solutions. Thus, for many interesting and potentially practical problems, computing {\em all} solutions in a reasonable time is a task beyond the reach of current symbolic-numeric and homotopy continuation solving methods.

\section{Examples of problem-solution manifolds worked out in detail}
\label{sec:PS-M-Examples}

Let us now provide a detailed explanation of the problem-solution manifold concept introduced in Sec.~\ref{sec:Problem-solution manifold} for the 5pt and 4pt problems solved in this work. 

\begin{example}
\label{ex:5p-SM}
Consider the 5pt problem of computing the relative pose of two calibrated cameras from 5 correspondences in two images,i.e., two $3\times 4$ matrices $C_1=\begin{bmatrix} R_1 & t_1 \end{bmatrix}=\begin{bmatrix} I & 0 \end{bmatrix},$ $C_2= \begin{bmatrix} R_2 & {t_2} \end{bmatrix}$ in the special Euclidean group $\SE_\RR (3)$, which view five world points ${X_1},\ldots , {X_5} \in \RR^3$ where $X_1$ is normalized to lie in the first image plane: $\{ {X_1} \mid {X}_1^{(3)} = 1 \} \cong \RR^2.$
The points are in front of both cameras iff their depths 
$$\lambda_{i,j} = \lambda_{i,j} (X , C) = {R_j}^{(3, :)} {X_i} + t_{j}^{(3)}$$ are all positive (${R_j}^{(3, :)}$ is the third row of $R_j$ and $t_{j}^{(3)}$ the third entry of ${t_i}$.)
Consider
\[
\arraycolsep=2pt
\begin{array}{cccccc}
\Psifpp: &\left(\RR^2 \times (\RR^3)^4\right) &\times& \SE_\RR (3)& 
\to& 
 \left(\RR^2\right)^{10} \times \RR^{9}\\
&(X &,& C )& 
\mapsto& (x ,  \lambda)
\end{array}
\]
where
\begin{equation}
\label{eq:img-pt}
x = \left(\lambda_{i,j}^{-1} \left(R_j {X_i} + {t_j}\right)^{(1:2)} \right)_{i=1,\ldots , 5;\ j=1,2;\ (i,j) \ne (1,1)}.
\end{equation}
Here the problem space is $P = \left(\RR^2\right)^{10}$
the solution space is $\calS=\RR^{9}$, $\pi(x, \lambda ) = x$, and
our problem-solution manifold $M=\Mfpp $
is the set of smooth points in the semialgebraic set $\im (\Psifpp) \cap \left(\left(\RR_{>0}\right)^{9} \times \left(\RR^2\right)^{10}\right).$
\end{example}
\begin{remark}
For a generic problem $x \in \Mfpp,$ the fiber $\pi^{-1} (x)$ consists of at most $10$ solutions $\lambda > 0$, and every such $\lambda$ can be extended uniquely to a pair $(X,C) \mapsto (x, \lambda ).$
\end{remark}
\begin{remark}
Our assumptions in~\cref{ex:5p-SM} imply that $\lambda_{1,1} =1.$
In subsequent sections, we treat the five-point problem as a system of equations in the nine remaining unknown depths.
More generally, we could dehomogenize our system by setting any linear form in $\lambda$'s equal to $1.$
\end{remark}

\begin{example}
\label{ex:Scr-SM}
Consider the Scranton relaxation of the 4pt problem computing the relative pose of three calibrated cameras from 4 correspondences in 3 images. We have 4 world points ${X_1},\ldots , {X_4}$ with ${X}_1^{(3)} = 1 $ and three cameras $C_1=\begin{bmatrix} I & 0 \end{bmatrix},$ $C_2= \begin{bmatrix} R_2 & {t_2} \end{bmatrix},$ $C_3= \begin{bmatrix} R_3 & {t_3} \end{bmatrix}$ such that cameras $C_1,C_2,C_3$ view $X_2,\ldots , X_4,$ cameras $C_2, C_3$ view $X_1,$ and camera $C_1$ views the line $\ell$ in the direction $e_2 = \begin{bmatrix} 0 & 1 & 0\end{bmatrix}^\top$ that passes through $X_1,$ parametrized as $\ell (l) = X_1 + l e_2.$  
Now, we have a map
\[
\arraycolsep=2pt
\begin{array}{cccccc}
\PsiScr: &\left(\RR^2 \times (\RR^3)^3 \times \RR \right) &\times& (\SE_\RR (3))^2& 
\to& 
 \left(\RR^2\right)^{12} \times \RR^{12}\\
&((X, l) &,& C )& 
\mapsto& \left(x ,  (\lambda, l) \right)
\end{array}
\]
where $x_{i,j}$ are as in~\cref{eq:img-pt} except that 
\[
x_{1,1} = \left({X_i} + {t_j} - l e_2\right)^{(1:2)} .
\]
Here the problem space is $P = \left(\RR^2\right)^{12}$
the solution space is $\calS=\RR^{12}$, $\pi(x, \lambda ) = x$, and
our problem-solution manifold $M=\MScr $
is the set of smooth points in the semialgebraic set $\im (\PsiScr) \cap \left(\left(\RR_{>0}\right)^{11} \times \RR \times \left(\RR^2\right)^{12}\right).$
\end{example}

\section{Additional details for 5pt formulation}
\label{sec:5pt-formulation-details}

Here we provide additional details concerning the solutions of the depth-formulated 5pt problem developed in~\Cref{sec:5pt}.
Given $(x,\lambda ) \in \Mfpp ,$ we first note that a valid rotation matrix $R (x, \lambda )$ may be estimated by computing certain auxiliary quantites: for $i \in \{ 1, \ldots , 5\}$ and $v\in \{ 1 , 2\}, $ we define $X_{i,v} = \lambda_{i,v} x _{i, v},$ and, for
distinct $i,j,k \in \{ 2, 3,4, 5 \},$
\begin{align*}
A_{i,j,k}^{(v)} &= \left(\begin{array}{c|c|c}
X_{i,v} - X_{1,v} &
X_{j,v} - X_{1,v} &
X_{k,v} - X_{1,v}
\end{array}\right).
\end{align*}
Thus, $\det A_{i,j,k}^{(v)}$ gives the oriented volume of a tetrahedron whose vertices are $X_{1,v},X_{i,v},X_{j,v},X_{k,v}.$
To estimate the rotation from a geometrically meaningful solution, one may compute either
\begin{equation}
\label{eq:Rot234}
R_2(x,\lambda ) = A_{2,3,4}^{(2)} \left(
A_{2,3,4}^{(1)}
\right)^{-1}
\end{equation}
or 
\begin{equation}
\label{eq:Rot235}
R_2(x,\lambda ) = A_{2,3,5}^{(2)} \left(
A_{2,3,5}^{(1)}
\right)^{-1}.
\end{equation}

The solutions to~\Cref{eq:5pt-equations} need not satisfy the additional constraint $\det R_2(x,\lambda ) =1$, since there is a sign-symmetry $\lambda_{i,v} \mapsto (-1)^{v+1} \lambda_{i, v}$ which changes the sign of $\det R_2 (x, \lambda)$ but leaves~\Cref{eq:5pt-equations} invariant.
Moreover, there are $76$ nonsingular, spurious solutions to the square subsystem obtained by dropping one equation, plus an additional $2$ of higher multiplicity.
For these $78$ spurious solutions, either of the matrices in~\cref{eq:Rot234} or~\cref{eq:Rot235} may have determinant $-1.$
These spurious solutions may be ruled out by enforcing $\det = 1$ for  either both of these matrices, or for one of these matrices in addition to all original depth constraints. 
With these constraints enforced, the generic number of solutions drops to $20.$
Moreover, there is an additional symmetry given by the ``twisted pair''~\cite{HZ-2003}: letting
\[
t (x, \lambda ) = \lambda_{1,2} v_{1,2} - R(x,\lambda ) \lambda_{1,1} v_{1,1},
\]
we define $tw (x, \lambda )$ coordinate-wise fixing $x$ and 
\[
\textrm{tw} (\lambda_{i,j}) =  \displaystyle\frac{(-1)^{j+1} \, \lVert t(x, \lambda ) \rVert^2 \,  \lambda_{i,j}} {\lVert \lambda_{i,2} v_{i,2} \rVert^2 - \lVert \lambda_{i,1} v_{i,1} \rVert^2}
\]
The map on p-s pairs $(x, \lambda ) \mapsto (x, \textrm{tw} (x, \lambda ))$
reverses the signs of depths in the second view. 
This justifies our claim that there are at most $10$ geometrically meaningful solutions to~\cref{eq:5pt-equations}.
We remark that the partition of $20$ solutions into twisted pairs is preserved along non-singular solution curves computed by our HC method.

\section{Additional details for Scranton formulation}
\label{sec:Scranton-formulation-details}

Unlike the system used for the 5pt problem, the square system for Scranton  has infinitely many solutions.
Recall that this system is given by the relaxed depth constraint
\begin{equation*}
    || v_{1,1} + l\,[0;1;0] - \lambda_{m,1} v_{m,1} ||^2 
  = || \lambda_{1,2} v_{1,2} - \lambda_{m,2} v_{m,2} ||^2
\end{equation*}
and~\cref{eq:4pt-equations} for remaining points and cameras.
A one-dimensional family of solutions may be obtained by setting all depths except $\lambda_{1,1}, \lambda_{1,2}, \lambda_{1,3}$ to $0,$ resulting in $2$ nontrivial equations in the remaining $3$ unknowns: namely,
\begin{align*}
    || v_{1,1} + l\,[0;1;0] ||^2 
  = || \lambda_{1,2} v_{1,2}  ||^2 \\
  || \lambda_{1,2} v_{1,2} ||^2 
  = || \lambda_{1,3} v_{1,3}  ||^2 
\end{align*}

The square system for Scranton also has several families of isolated singular solutions where certain depths equal $0.$
However, for generic data, the number of \emph{nonsingular} solutions equals the number of solutions with \emph{nonzero depths}, which is $1408.$
Among these, there is a four-fold sign symmetry where $\lambda_{i,2} \mapsto \pm \lambda_{i,2}, \lambda_{i,3} \mapsto \pm \lambda_{i,3},$ and $320 = 4 \times 80$ cannot be lifted to a valid pair of rotations $\left(R_2 (x, \lambda) , R_3 (x, \lambda ) \right).$

Taking these facts into account, there are at most $1408 - 3\times 272 - 4 \times 80 = 272$ geometrically relevant solutions on the problem-solution manifold.
This agrees with the number of solutions reported in both~\cite{Kileel-MPCTV-2016} and~\cite{PL1P}, where formulations in terms of trifocal tensors and camera matrices, respectively, were employed.
We note that, unlike the five-point problem, there is no further reduction in the number of solutions implied by a symmetry such as the twisted pair; this follows by numerically computing the Galois group associated to Scranton, using  techniques described in~\cite{duff2021galois}, which turns out to be the symmetric group on $272$ letters.

\begin{figure*}
    \centering
    \includegraphics[width=\linewidth]{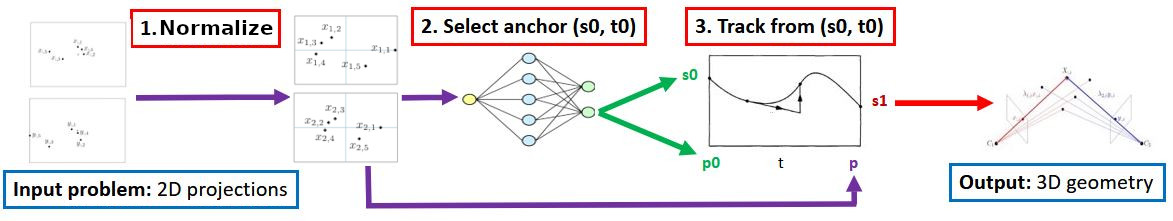}
    \caption{Illustration of testing the classifier. A minimal sample $p$ is obtained in the RANSAC scheme. Then, the sample is normalized. The normalized sample is used as the input to the trained MLP, which selects the starting p-s pair $(p_0, s_0)$. Then, HC is tracked from $(p_0,s_0)$ to $p$. If a solution $s$ is obtained, it is converted to a relative pose and the RANSAC score of it is evaluated. }
    \label{fig:method}
\end{figure*}

\section{Additional details on HC methods}
\label{sec:HC-details}
As noted in~\cref{sec:hc_solving}, our homotopy $H$ depends on the choice of a path $p(t)$ connecting $p_0$ to $p,$ where $(p_0, s_0)$ is a known p-s pair and $p$ is the problem to be solved. 
In all of our experiments, we consider one of two choices.
Mostly, we use {\bf 1) Linear segment HC}: that is, we choose 
$p(t) = (1-t)\,p_0 + t\,p$. 
 This linear segment homotopy has several advantages; among them, the straight-line programs needed to evaluate $H$ and its derivatives are much simpler than for other paths, and the fact that $p(t)$ is real-valued for all $t.$ 
 However, under Linear segment HC, a differentiable solution curve $s(t)$ satisfying $H(p(t), s(t)) = 0$ need not exist for all $t \in [0,1]$. 
 For instance, a problem with singular solutions may exist somewhere along the segment connected in $P$ connecting $p_0$ and $p.$
 However, the solution curve $s(t)$ will exist for all $t\in [0,1]$ if $p_0$ and $p$ are ``close enough"---more precisely, if $p(t)$ avoids the lower-dimensional set of critical values of $\pi$ in $P$ for all $t\in [0,1].$
 
Alternatively, one may consider \noindent {\bf 2) Circular arc HC}: here, we reparametrize the segment $p(t)$ via a circular arc $t (\tau): [0,1] \to [0,1]$ obtained by fixing a random $\gamma \in \CC $ (typically of modulus $1$) and  $t (\tau ) = \frac{\gamma \, \tau }{1 + (\gamma -1 ) \tau}$. This is the $\gamma $-trick of~\cite[Lemma 7.1.3 ]{DBLP:books/daglib/0014410}.
Numerical continuation with the resulting homotopy $H(s,\tau )$ is \emph{globally convergent with probability one}: for almost all choices of $\gamma ,$ the solution curves $s(t)$ are defined for all $t\in [0,1],$ and any isolated target solution $(p,s)$ is the endpoint of some solution curve.
However, this necessitates computing many spurious solutions.
An experimental comparison of Linear segment and Circular arc HC may be found in~\cref{tab:ablation-hc-study}.

We now explain why the square systems used in our homotopies are sufficient when starting from a p-s pair on the problem-solution manifold.
Consider a path $t \mapsto (p(t), s(t)) \in P \times \calS$ where $(p(0), s(0)) = (p_0, s_0)\in M$ and such that that the $n\times n$ Jacobian matrix $\frac{d \, H}{d \, s} (p(t), s(t))$ has rank $n$ for all $t\in [0,1].$
Thus, the path $t\mapsto (p(t), s(t))$ is contained in a single connected component of the set of nonsingular points in the complex vanishing set $\{ (p, s) \in P_\CC \times \calS_\CC \mid f(p, s) = 0 \}.$
Among these connected components is the set of smooth points in the Zariski closure of $M$. 
Indeed, the complex Zariski closure of $M$ has a rational parametrization (given by one of the maps $\Psifpp, \PsiScr$ defined in~\cref{sec:PS-M-Examples}), so it is irreducible, and the connectedness of its smooth points follows by~\cite[pp.~21--22]{GriffHarr}.
Since the point $(p_0, s_0)$ is contained in this connected component, so also must $(p(t), s(t))$ for all $t\in [0,1].$
Thus, any polynomial $g(p,s)$ vanishing on $M$ satisfies $g(p(t),s(t))=0$ for all $t\in [0,1].$ This means that, if HC tracking from $(p_0, s_0) \in M$ succeeds using our square system of constraints, then \emph{all} additional constraints which are polynomial equalities are automatically satisfied.
This justifies the fact that we do not explicitly enforce constraints like $\det R_2(x , \lambda ) =1$, since it is enough to enforce them for the initial p-s pair.

\subsection{Efficient evaluation of predictor/corrector}\label{sec:eff_eval}

The Runge-Kutta method used for the predictor step and Newton's method used for the corrector step in our HC implementation both require solving systems of linear equations.
In either step, the coefficient matrix is given by the Jacobian $\frac{\partial H(s,t)}{\partial s}$  (Sec.~{\ref{sec:hc_solving}}). 
In the case of the depth formulation of the Five-Point problem \eqref{eq:5pt-equations} and the Four-Point problem \eqref{eq:4pt-equations}, the associated Jacobian matrix is sparse. The sparsity pattern of the Jacobian matrix $\frac{\partial H(s,t)}{\partial s}$ is shown in \eqref{eq:sparse} for the Five-Point problem, and in \eqref{eq:sparse4} for the Four-Point problem.

\hide{
\tim{Does this look better? not really....}
\begin{equation}
    \begin{bmatrix}
    \ast & 0 & 0 & 0 & \ast & \ast & 0 & 0 & 0 \\
    0 & \ast  & 0 & 0 & \ast & 0 & \ast  & 0 & 0 \\
    \ast & \ast  & 0 & \ast & 0 & 0 & \ast & 0 & 0 \\
    0 & 0 & \ast & 0 & \ast & 0 & 0 & \ast & 0 \\
   \ast  & 0 & \ast & 0 & 0 &\ast  & 0 & \ast & 0 \\
    0 & \ast & \ast  & 0 & 0 & 0 & \ast  & \ast & 0 \\
    0 & 0 & 0 & \ast & \ast & 0 & 0 & 0 & \ast \\
   \ast & 0 & 0 & \ast & 0 & \ast & 0 & 0 & \ast \\
    0 & \ast & 0 & \ast & 0 & 0 & \ast & 0 & \ast \\
    \end{bmatrix}\label{eq:sparse}
\end{equation}
}

\begin{equation}
    \begin{bsmallmatrix}
    A_{0,3} & 0 & 0 & 0 & A_{0,4} & A_{0,5} & 0 & 0 & 0 \\
    0 & A_{1,1} & 0 & 0 & A_{1,4} & 0 & A_{1,6} & 0 & 0 \\
    A_{2,0} & A_{2,1} & 0 & A_{2,3} & 0 & 0 & A_{2,6} & 0 & 0 \\
    0 & 0 & A_{3,2} & 0 & A_{3,4} & 0 & 0 & A_{3,7} & 0 \\
    A_{4,0} & 0 & A_{4,2} & 0 & 0 & A_{4,5} & 0 & A_{4,7} & 0 \\
    0 & A_{5,1} & A_{5,2} & 0 & 0 & 0 & A_{5,6} & A_{5,7} & 0 \\
    0 & 0 & 0 & A_{6,3} & A_{6,4} & 0 & 0 & 0 & A_{6,8} \\
    A_{7,0} & 0 & 0 & A_{7,3} & 0 & A_{7,5} & 0 & 0 & A_{7,8} \\
    0 & A_{8,1} & 0 & A_{8,3} & 0 & 0 & A_{8,6} & 0 & A_{8,8} \\
    \end{bsmallmatrix}\label{eq:sparse}
\end{equation}


\begin{equation}
{\tiny 
    \begin{bsmallmatrix}
    A_{0,0} & 0 & 0 & A_{0,3} & A_{0,4} & 0 & 0 & 0 & 0 & 0 & 0 & 0\\
    0 & A_{1,1} & 0 & A_{1,3} & 0 & A_{1,5} & 0 & 0 & 0 & 0 & 0 & 0\\
    A_{2,0} & A_{2,1} & 0 & 0 & A_{2,4} & A_{2,5} & 0 & 0 & 0 & 0 & 0 & 0\\
    0 & 0 & A_{3,2} & A_{3,3} & 0 & 0 & A_{3,6} & 0 & 0 & 0 & 0 & 0\\
    A_{4,0} & 0 & A_{4,2} & 0 & A_{4,4} & 0 & A_{4,6} & 0 & 0 & 0 & 0 & 0\\
    0 & A_{5,1} & A_{5,2} & 0 & 0 & A_{5,5} & A_{5,6} & 0 & 0 & 0 & 0 & 0\\
    A_{6,0} & 0 & 0 & 0 & 0 & 0 & 0 & A_{6,7} & A_{6,8} & 0 & 0 & 0\\
    0 & A_{7,1} & 0 & 0 & 0 & 0 & 0 & A_{7,7} & 0 & A_{7,9} & 0 & 0\\
    A_{8,0} & A_{8,1} & 0 & 0 & 0 & 0 & 0 & 0 & A_{8,8} & A_{8,9} & 0 & 0\\
    0 & 0 & A_{9,2} & 0 & 0 & 0 & 0 & A_{9,7} & 0 & 0 & A_{9,10} & A_{9,11}\\
    A_{10,0} & 0 & A_{10,2} & 0 & 0 & 0 & 0 & 0 & A_{10,8} & 0 & A_{10,10} & A_{10,11}\\
    0 & A_{11,1} & A_{11,2} & 0 & 0 & 0 & 0 & 0 & 0 & A_{11,9} & A_{11,10} & A_{11,11}
    \end{bsmallmatrix}}
    \label{eq:sparse4}
\end{equation}

Using generic methods such as LU decomposition, as in previous work \cite{Fabbri-PAMI-2020}, numerical linear algebra becomes a significant bottleneck in both the predictor and corrector stages.
To overcome this bottleneck, we replace the generic numerical linear algebra with closed-form solutions to the systems of linear equations with coefficient matrices \eqref{eq:sparse} and \eqref{eq:sparse4}. 
This replacement results in about 5x speedup for both problems.


\section{Variations of the normalization}
\label{supp:normalizations}
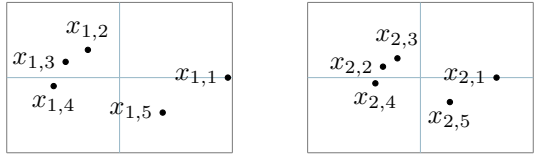
\begin{figure}
    \centering
    
    \begin{tikzpicture}[scale = 1]
    
    \draw[gray] (-1.5,-1) -- (-1.5,1);
    \draw[gray] (1.5,-1) -- (1.5,1);
    \draw[gray] (-1.5,-1) -- (1.5,-1);
    \draw[gray] (-1.5,1) -- (1.5,1);
    \draw[cyan!30!gray!70] (0,-1) -- (0,1);
    \draw[cyan!30!gray!70] (-1.5,0) -- (1.5,0);
    
    \filldraw[black] (1.44, 0) circle (1pt) node[anchor=east] {$x_{1,1}$};
    \filldraw[black] (-0.421, 0.369) circle (1pt) node[anchor=south] {$x_{1,2}$};
    \filldraw[black] (-0.717, 0.2095) circle (1pt) node[anchor=east] {$x_{1,3}$};
    \filldraw[black] (-0.876, -0.113) circle (1pt) node[anchor=north] {$x_{1,4}$};
    \filldraw[black] (0.5735, -0.4655) circle (1pt) node[anchor=east]     {$x_{1,5}$};

    \draw[gray] (2.5,-1) -- (2.5,1);
    \draw[gray] (5.5,-1) -- (5.5,1);
    \draw[gray] (2.5,-1) -- (5.5,-1);
    \draw[gray] (2.5,1) -- (5.5,1);
    \draw[cyan!30!gray!70] (4,-1) -- (4,1);
    \draw[cyan!30!gray!70] (5.5,0) -- (2.5,0);

    \filldraw[black] (5.0116, 0) circle (1pt) node[anchor=east] {$x_{2,1}$};
    \filldraw[black] (3.69395, 0.25625) circle (1pt) node[anchor=south] {$x_{2,3}$};
    \filldraw[black] (3.502, 0.14545) circle (1pt) node[anchor=east] {$x_{2,2}$};
    \filldraw[black] (3.4009, -0.07625) circle (1pt) node[anchor=north] {$x_{2,4}$};
    \filldraw[black] (4.3916, -0.32545) circle (1pt) node[anchor=north] {$x_{2,5}$};
    
    \end{tikzpicture}
    \caption{An example of a 5pt problem after normalization of image coordinates.}
    \label{fig:normalized-5pt}
\end{figure}

\hide{The general strategy for the normalization of the 5pt problem and the Scranton problem goes as follows:
\begin{enumerate}
    \item Rotate the center of mass of the projections in all views to zero.\\(OR)\\
    Rotate the point closest to the center of mass to zero.
    \item Take the point and view for which the projection is farthest from zero.
    \item Order the views by the distance of the point obtained in 2. to zero.
    \item Rotate the point obtained in 2. to the $x$-axis.\\(OR)\\Rotate the direction of the largest variance to the $x$-axis such, that the point from 2. is on the right.
    \item Order the projections CCW in the first view.
\end{enumerate}
\noindent
We considered the following variants of the general strategy:
\begin{enumerate}[A.]
    \item Rotate the center of mass to zero, rotate the point farthest from zero to $x$-axis.
    \item Rotate the closest point to center of mass to zero and the point farthest from zero to $x$-axis.
    \item Rotate the center of mass to zero and the maximal variance to $x$-axis.
    \item Rotate the closest point to center of mass to zero and the maximal variance to $x$-axis.
\end{enumerate}
}

Let us provide additional details about our normalization of problems to simplify their variability and thus to make learning of the picking function $\sigma$ easier. 

Fig.~\ref{fig:normalized-5pt} shows an example of the normalized 5pt problem. The mean direction vectors $m_1, m_2$ in both images are at $[0;0;1]$. The first correspondence, $x_{1,1}, x_{2,1}$ is on the $x$ axis. The first image is chosen such that it contains the larger angle with its corresponding $m_u$. This also means that the first correspondence point has a larger $x$ image coordinate: $x_{1,1}^{(1)}>, x_{2,1}^{(1)}$. To make the normalized problems independent on the ordering of the correspondences, we sort them by their polar angles in the coordinate system in the the first image. Notice that their order may be swapped in the second image, e.g., as for $x_{2,2}$, $x_{3,2}$. Such a swap is mainly due to a large change of the order of depth of the corresponding points in the scene, as seen from different view points, which is in practice much less frequent than keeping the order~\ifArXiV[70]\fi. 

Our normalization is chosen as the best one among several meaningful alternatives. The evaluation of the alternative normalization methods is shown in Tab.~\ref{tab:inv5} for the 5pt problem and in Tab.~\ref{tab:inv4} for the Scranton problem. The tables show that our strategy, labeled by A, which rotates the center of mass to zero and the farthest point on the $x$-axis, has the best success rate for both problems. Note that every normalization strategy performs better than when tracking without normalization. 

The normalization strategies are as follows:
\begin{enumerate}[A.,itemsep=1pt,parsep=1pt]
    \item Rotate the center of mass to zero, rotate the point farthest from zero to $x$-axis.
    \item Rotate the center of mass to zero with an iterative procedure, rotate the point farthest from zero to $x$-axis.
    \item Rotate the closest point to center of mass to zero and the point farthest from zero to $x$-axis.
    \item Rotate the center of mass to zero and the maximal variance to $x$-axis.
    \item Rotate the closest point to center of mass to zero and the maximal variance to $x$-axis.
\end{enumerate}

In the case of the Scranton problem, we also have to decide which point in which view to relax on the line. Here, we consider:
\begin{enumerate}[a),itemsep=1pt,parsep=1pt]
    \item The farthest point and view.
    \item The point rotated to zero (if possible).
\end{enumerate}

Our normalization induces three linear constraints for every view. The instance $p$ of the 5pt problem consists of 2D projections of 5 points into two views, therefore $p \in \RR^{20}$. The normalized instances live in a $20-2 \times 3 = 14$ dimensional subspace of $\RR^{20}$. The instance $p$ of the 4pt problem consists of 2D projections of 4 points into 3 views, $p \in \RR^{24}$. The normalized instances live in a $24-3 \times 3 = 15$ dimensional subspace of $\RR^{24}$.

The values of the success rate in this experiment are low because we use randomly sampled data and track from every p-s pair to every other p-s pair. Tab.~\ref{tab:HC-results} shows that the success rate significantly increases if we track from preselected anchors which are chosen to perform well and we select the best starting anchor with a trained classifier.

\begin{table}
    \centering
    \begin{tabular}{|l||c|c|c|}
    \hline
        Strategy & Succ. rate & Time inv. & Time HC \\
        \hline \hline
        No inv. & 0.53\% & 0 & 14.25 $\mu s$ \\ \hline
        A. & \textbf{3.73\%} & \textbf{0.43} $\mu s$ & 11.80 $\mu s$\\ \hline
        B. & 3.70\% & 0.77 $\mu s$ & 11.88 $\mu s$ \\ \hline
        C. & 2.29\% & 0.36 $\mu s$ & 11.19 $\mu s$\\ \hline
        D. & 1.56\% & 1.03 $\mu s$ & 12.18 $\mu s$\\ \hline
        E. & 0.71\% & 0.68 $\mu s$ & 10.44 $\mu s$\\ \hline
    \end{tabular}
    \caption{Evaluation of the normalization for the 5 pt problem. We have generated $4000$ problem-solution pairs, normalized them with a given strategy and tracked HC from every p-s pair to every other. We consider strategies from Sec.~\ref{supp:normalizations}. We measure the success rate, average time of the normalization and of HC. The track is considered successful if the squared Euclidean distance from the obtained solution to the ground-truth is less than $10^{-5}$.}
    \label{tab:inv5}
\end{table}

\begin{table}
    \centering
    \begin{tabular}{|l||c|c|c|c|}
        \hline
        Strategy & Line strategy & Succ. rate & Time inv. & Time HC \\
        \hline \hline
        No inv. & - & 0.21\% & 0 & 22.11 $\mu s$ \\ \hline
        A. & a) & \textbf{1.44\%} & \textbf{0.50 $\mu s$} & 17.82 $\mu s$ \\ \hline
        B. & a) & \textbf{1.44\%} & 1.16 $\mu s$ & 17.40 $\mu s$ \\ \hline
        C. & a) & 0.80\% & 0.82 $\mu s$ & 19.57 $\mu s$ \\ \hline
        C. & b) & 0.32\% & 0.78 $\mu s$ & 17.37 $\mu s$ \\ \hline
        D. & a) & 0.61\% & 1.39 $\mu s$ & 20.13 $\mu s$ \\ \hline
        E. & a) & 0.37\% & 1.01 $\mu s$ & 20.43 $\mu s$ \\ \hline
        E. & b) & 0.27\% & 0.93 $\mu s$ & 18.46 $\mu s$ \\ \hline
    \end{tabular}
    \caption{Evaluation of the normalization for the Scranton problem. We have generated $4000$ problem-solution pairs, normalized them with a given strategy and tracked HC from every p-s pair to every other. We consider strategies from Sec.~\ref{supp:normalizations}. We measure the success rate, average time of the normalization and of HC. The track is considered successful if the squared Euclidean distance from the obtained solution to the ground-truth is less than $10^{-5}$.}
    \label{tab:inv4}
\end{table}

\section{Study to justify engineering choices}
\label{sec:ablations}
Let us describe the data sets we use to study our engineering choices.

Training data set $\Dfp$ consists of 40000 p-s pairs. We randomly sample pairs of cameras and 5-tuples of 3D points from the ETH 3D dataset~\cite{schoeps2017cvpr} ``Office'' and ``Terrains''. Problem parameters $p$ are 10D vectors of 2D image coordinates obtained by projecting the sampled 5-tuples of 3D points by the camera pairs. The corresponding solutions $s$ are 10D vectors of the depths of the 3D points in the two cameras normalized to have the first depth equal to 1. Data set $\DScr$, consisting of 40000 p-s pairs, is constructed analagously by sampling 4-tuples of 3D points and triplets of cameras. Data sets are used to select anchor sets and to train our model for selecting the starting problem-solution pair. 

Anchor sets $\Afp_{50}, \Afp_{75}, \Afp_{90}, \Afp_{100}$ are selected by the procedure described in Sec.~\ref{sec:anchors} such that $\Afp_{50} \subset \Afp_{75} \subset \Afp_{90} \subset \Afp_{100} \subset \Dfp$ where $\Afp_{50}$ of $8$ anchors covers 50\% of problems in $\Dfp$, $\Afp_{75}$ of 26 anchors covers 75\% of problems in $\Dfp$, $\Afp_{90}$ of 70 anchors covers 90\% of problems in $\Dfp$ and $\Afp_{100}$ of 465 anchors covers 100\% of problems in $\Dfp$. Data sets, $\AScr_{50}$, of 16 anchors,  $\AScr_{75}$, of 50 anchors, $\AScr_{90}$, of $134$ anchors and $\AScr_{100}$, of 1205 anchors, are constructed analagously from $\AScr$.

Test data set $\Vfp$ consists of 60000 p-s pairs constructed as above from the ETH 3D dataset ``Delivery area'' and ``Facade''.  Data set $\VScr$, consisting of 60000 p-s pairs, is constructed analagously.  We use $\Vfp$ and $\VScr$ in the experiments studying anchor set selection methods reported in Tab.~\ref{tab:eval_mlp}.

\begin{table*}
\begin{center}
  \begin{tabular}{|l||r|r|r|}
      \hline
      & \multicolumn{3}{c|}{Succ. rate $\mu_s \pm \delta_s$ [\%] / Time $\mu_t$ [$\mu s$]}\\ \hline
      & \multicolumn{3}{c||}{5pt problem} \\      
      \hline
      Solving technique    & M2~\cite{leykin2011numerical} & MINUS~\cite{TRPLP} & OUR   \\ \hline \hline
      $\CC$-HC, All Sols   & 98.9
$\pm$ 0.2/ \SI{1.9 E 5} &  97.7 $\pm$ 0.2 / 15197.1 & 97.7 $\pm$ 0.1 / 5133.7\\ \hline
      $\CC$-HC, $\RR$ Sols & 56.1 $\pm$ 3.3  / \SI{1.2 E 5} & 55.3 $\pm$ 2.6 / 5704.7  & 54.7 $\pm$ 3.2 / 1895.1  \\ \hline
      $\CC$-HC, Fab Sol    &  $12.9 \pm 0.9$ /  \SI{9.8 E 4}&   12.0 $\pm$ 1.5 / 638.0 & 13.1 $\pm$ 1.4 / 165.8  \\ \hline
      $\RR$-HC, $\RR$ Sols & 9.9 $\pm$ 1.7 / \SI{5.7 E 4} & 9.7 $\pm$ 1.5 / 647.9  & 9.7 $\pm$ 1.5 / 106.8  \\ \hline
      $\RR$-HC, Fab Sol    &  3.4$\pm$ 1.3 / \SI{4.4 E 4} & 2.7 $\pm$ 0.8 / 67.6  & 2.7 $\pm$ 0.8 / 11.1  \\ \hline
      Newton, Fab Sol     & 4.0$\pm $ 0.6    / \SI{1.4 E 4}  &  4.0 $\pm$ 0.7 / 8.5  & 4.0 $\pm$ 0.7 / 1.3   \\ \hline
  \end{tabular}
  \\[1em]
    \begin{tabular}{|l||r|r|}
      \hline
      & \multicolumn{2}{c|}{Succ. rate $\mu_s \pm \delta_s$ [\%] / Time $\mu_t$ [$\mu s$]}\\ \hline
      &  \multicolumn{2}{c|}{Scranton}\\      
      \hline
      Solving technique    &  MINUS~\cite{TRPLP} & OUR   \\ \hline \hline
      $\CC$-HC, All Sols   &  95.5 $\pm$ 2.6 / 608332.0 & 95.7 $\pm$ 2.6 / 187364.5  \\ \hline
      $\CC$-HC, $\RR$ Sols & 22.7 $\pm$ 1.6 / 47961.4 & 22.5 $\pm$ 1.7 / 14905.1  \\ \hline
      $\CC$-HC, Fab Sol    &  3.5 $\pm$ 0.7 / 1280.7  & 3.3 $\pm$ 0.9 / 405.7  \\ \hline
      $\RR$-HC, $\RR$ Sols &  7.5 $\pm$ 1.2 / 2548.4 & 7.6 $\pm$ 1.2 / 484.4  \\ \hline
      $\RR$-HC, Fab Sol    &  1.2 $\pm$ 0.3 / 70.2  &  1.2 $\pm$ 0.3 / 14.1 \\ \hline
      Newton, Fab Sol     &   1.9 $\pm$ 0.6 / 8.6  & 2.0 $\pm$ 0.4 / 1.4  \\ \hline
  \end{tabular}
  \end{center}
  \caption{Homotopy continuation study. The rows represent variations mixing the solving technique of complex ($\CC$-HC) and real ($\RR$-HC) homotopy continuation, the Newton's local method (Newton) with starting from all solutions (All Sols), real solutions only ($\RR$ Sols), and the fabricated solution only (Fab Sol) of a problem. The columns represent different implementations of homotopy continuation. M2 denotes the off-the-shelf implementation in Macaulay2~\cite{leykin2011numerical}. MINUS denotes the implementation based on~\cite{TRPLP}. OUR denotes our efficient implementation. To compare MINUS and OUR, we selected $20$ subsets $P_i$, $i=1,\ldots,20$, each containing $50$ random problems from $\Pfp$ for $5$pt problem and from $\PScr$ for Scranton.
  All problems were normalized. For each $P_i$, we compute the success rate $\mu_{s_i}$ of $50^2-50$ of homotopy continuations from each start problem $p_{ij} \in P_i$ to each different target problem $p_{ik} \in P_i$. We consider a homotopy continuation successful if the fabricated solution of the target problem $p_{ik}$ is among the solutions reached by the homotopy continuation within $10^{-5}$ Euclidean distance in the solution space of depths. We report the mean success rate $\mu_s = mean(\mu_{s_i})$ and the standard deviation $\delta_s = std(\mu_{s_i}$) over all $P_i$'s for each implementation and mean computation times $\mu_t$.}
  \label{tab:ablation-hc-study}
\end{table*}


\subsection{Comparison of different tracking approaches}
\label{subsec:ablation}

Data set $\Pfp$ consists of of 3751 p-s pairs sampled from the ETH 3D dataset ``Courtyard''. $\Afp$ and $\Pfp$ are disjoint. All problems in $\Pfp$ are checked to be generic and can be used as good starting problem-solution pairs. We select 20 random pairwise disjoint $\Pfp_i \subset \Pfp$ consisting of $50$ problem-solution pairs. Data sets $\PScr$, consisting of 5727 problem-solutions pairs, and $\PScr_i$, consisting of 50 p-s pairs, are constructed analagously. We use $\Pfp_i$'s and $\PScr_i$ in the experiments studying variations of the homotopy continuation methods reported in Tab.~\ref{tab:ablation-hc-study}. In this table, we measure the success rate for a given subset $P_i$ as a percentage of different pairs $p_{i,j} \in P_i, p_{i,k} \in P_i$ for which the fabricated solution to the target problem $p_{i,k}$ can be recovered when tracking from $p_{i,j}$ to $p_{i,k}$. Then, we find the mean success rate $\mu_s$, and the standard deviation $\delta_s$ over all subsets $P_i, i \in \{1,...,20\}$.

Tab.~\ref{tab:ablation-hc-study} shows that for every setting, our evaluation (Sec.~\ref{sec:eff_eval}) brings about 5x speedup over the previous work \cite{TRPLP} without any impact on the success rate.

The table also shows that Homotopy Continuation in the complex domain tracked from every solution has almost $100 \%$ success rate, but the running time of the solver is prohibitively slow to be used in the RANSAC scheme. We can see that reducing the number of tracks, as well as tracking in $\RR$ instead of $\CC$ can significantly reduce the running time (about 10000x for the Scranton problem) at the cost of a lower success rate. Note (Tab.~\ref{tab:HC-results}), that the issue with the low success rate may be remedied by selecting an appropriate starting problem-solution pair $(p_0, s_0)$ for every target problem $p$.

Tab.~\ref{tab:ablation-hc-study} also shows that using the Newton method may be a promising approach for solving the minimal problems, as it has a higher success rate and lower running time than Homotopy Continuation with the same setting. Therefore, we have found starting p-s pairs (Sec.~\ref{sec:anchors}) and trained the MLP for the Newton method. The comparison of the solvers using the Homotopy Continuation and the Newton method is shown in Tab.~\ref{tab:HC-vs-N}. We can see that the effective time (the average time needed to obtain one correct solution in the RANSAC scheme) of the solver using Homotopy Continuation is about 2x lower than the effective time of the solver using Newton method. While the solvers using Newton method are faster, the success rate of the solver, which combines Homotopy Continuation and MLP classifier has a higher success rate. The possible explanation for this is that the Newton method behaves ``more randomly'' than the Homotopy Continuation, and therefore, it is more difficult to train.

\begin{table}
\begin{center}
  \begin{tabular}{|l||c|c|c|}
    \hline
    Method    &$\rho$[\%]&$\mu_t$[$\mu s$]&$\epsilon_t$[$\mu s$]\\ \hline \hline
    N3 + B1 & 0.01 & 1.73 & 15159.7 \\ \hline
    N3 + MLP & 0.12 & 8.2 & 6811.2 \\ \hline
    N15 + B1 & 2.1 & 2.6 & 119.7 \\ \hline
    N15 + MLP & 10.6 & 10.6 & 100.0 \\ \hline
    HC + B1 & 4.2 & 18.7 & 442.1 \\ \hline
    HC + MLP & 26.3 & 16.2 & \textbf{61.6} \\ \hline
    
  \end{tabular}
\end{center}
\caption{Study of methods for starting problem selection and tracking for Scranton problem. The strategies are evaluated on datasets \textit{Deilvery\_area} and \textit{Facade}. Tracking methods: `N3': Newton method with 3 steps, `N15': Newton method with 15 steps (this number of steps maximizes the efficient time), `HC': Homotopy Continuation as described in Sec~\ref{sec:hc_solving}. Problem selection methods: `B1': Tracks always from the same anchor. `MLP': selecting the starting problem as the one with the highest score given by the MLP.}
\label{tab:HC-vs-N}
\end{table}

Tab.~\ref{tab:HC-results} provides a more detailed analysis of the frequency of different results of real Homotopy Continuation tracked from the fabricated solution. We consider two different settings. In the ``All pairs'' setting, we track from each $p_i \in P$ to each other $p_j \in P$. Then, in the MLP setting, we select a starting p-s pair $(p_0, s_0)$ from $A_{90}$, and track from $p_0$ to $p$. We measure how often we reach the fabricated solution, non-fabricated meaningful solution, a non-valid solution, and how often HC fails and does not deliver any solution. The table shows that the MLP increases the probability of reaching the fabricated solution about 10x for 5pt and 20x for Scranton. The probability of reaching another meaningful solution increases about 3x for both problems, while the probability of reaching a non-valid solution and the probability of failing decreases.

\begin{table}
\begin{center}
  \begin{tabular}{|l||c|c||c|c|}
    \hline
    & \multicolumn{2}{c||}{All pairs} & \multicolumn{2}{c|}{MLP}\\
    \hline
    Result [\%] & 5 pt & 4 pt & 5pt & 4pt\\  \hline \hline
    Fabricated sol. & 3.64  & 1.49 & 38.8 & 29.2\\
    1 rel. pose correct & - & 0.74 & - & 4.53 \\
    Other meaningful & 9.50 & 13.43 & 31.6 & 34.2 \\
    Sol. with det -1 & 0.00 & 0.00 & 0.00 & 0.00 \\
    Sol. with zeros & 0.03 & 1.53 & 0.01 & 0.52 \\
    Negative sol. & 0.52 & 10.86 & 0.83 & 8.54 \\
    Failed track & 86.31 & 71.94 & 28.8 & 23.0 \\
    \hline
  \end{tabular}
\end{center}
\caption{Percentage of different results of real homotopy continuation. In ``All pairs", we track from each $p_i \in P$ to each other $p_j \in P$. In ``MLP", we solve each $p \in V$ by selecting a starting p-s pair $(p_0, s_0)$ from $A_{90}$, and by tracking HC from $p_0$ to $p$. ``Fabricated sol." means that the fabricated solution was reached, ``1 rel. pose correct" means that if we convert the obtained solution of the Scranton problem to the relative poses, then at least one of three relative poses is correct. ``Other meaningful" means that a non-fabricated solution with positive depths and valid rotation matrix was reached, ``Sol. with det -1" means that the matrix $R$ is not a valid rotation. ``Sol. with zeros" means that some depths are equal to $0$, ``Negative sol." means that some depths are negative, and ``Failed track" means that the HC track has failed and, thus, the solution has not been found. }
\label{tab:HC-results}
\end{table}

\subsection{Comparison of different settings our solver}

Here, we show how different settings of the solver influence the resulting success rate $\rho$, mean running time $\mu_t$, and efficient time $\epsilon_t$. We perform this study on our solver for the Scranton problem. For every experiment, we use the settings from the main paper, except that one parameter is varied. 
The solver uses anchors $\AScr_{90}$ and it is evaluated on data $\VScr$. Our goal is to show that we have selected the optimal settings for our solver.

We compare different methods of dehomogenizing Scranton problem in Tab.~\ref{tab:ablation-formulation-study}. 
This justifies our choice of fixing the first depth $\lambda_{1,1}=1$, which gives a superior success rate compared to ``symmetric" and  ``asymmetric" dehomogenization proposed in \cite{QuanTM2006}.

See Fig.~\ref{fig:MLP_code} for a code snippet showing the structure of our MLP model. In Tab.~\ref{tab:MLP-size-study}, we show how the success rate $\rho$ and running time $\mu_t$ depends on the size of the MLP that is used for the classification of the anchors. 
Larger networks have a higher success rate. However, the efficient time of the solver with the smaller MLP is better because the time needed for the evaluation of the MLP grows faster than the success rate.

Finally, in Tab.~\ref{tab:track-numbers-study}, we show how the number of HC tracks per problem influences the success rate $\rho$, running time $\mu_t$, and efficient time $\epsilon_t$. In this experiment, we use the MLP trained in \ref{sec:start-p-s}, and we perform $n$ tracks from $n$ anchors with the highest score. We consider the solution successful if at least one track reaches the fabricated solution of the target problem. Much like using larger MLP as Tab.~\ref{tab:MLP-size-study}, such a strategy involving multiple anchors suggests a future approach to improving our solvers' success rates. 
However, we note that the efficient time $\epsilon_t$ in~\cref{tab:track-numbers-study} grows with the number of tracks, since the evaluation time grows faster than the success rate.

\begin{table}
\begin{center}
  \begin{tabular}{|l||c|}
    \hline
    Formulation & Succ. rate\\  \hline \hline
    First depth fixed & \textbf{2.30 \%} \\ \hline
    Quan symmetrical \cite{QuanTM2006} & 1.47 \% \\ \hline
    Quan asymmetrical \cite{QuanTM2006} & 1.13 \% \\ \hline
  \end{tabular}
\end{center}
\caption{Scranton dehomogenization study. Rows correspond to different formulations of the problems. For each method, we compute the success rate of $4000^2-4000$ HC calls from each starting p-s pair to each other target p-s pair. We consider the result successful if the fabricated solution of the target problem is and the result computed by HC are sufficiently close ( $\le 10^{-5}$ Euclidean distance  in the solution space of depths.)}
\label{tab:ablation-formulation-study}
\end{table}

\begin{table}
\begin{center}
  \begin{tabular}{|l||c|c|c|}
    \hline
    Layer size    &$\rho$[\%]&$\mu_t$[$\mu s$]&$\epsilon_t$[$\mu s$]\\ \hline \hline
    100 & 27.8 & 20.3 & \textbf{73.1} \\ \hline
    200 & 31.3 & 30.8 & 98.3 \\ \hline
    500 & 34.7 & 79.0 & 227.6 \\ \hline
    
  \end{tabular}
\end{center}
\caption{Study of different MLP sizes. The strategies are evaluated on datasets \textit{Deilvery\_area} and \textit{Facade}. Scranton problem, MLP+HC, $A_{75}$. Rows correspond to different sizes of hidden MLP layers.}
\label{tab:MLP-size-study}
\end{table}

\begin{table}
\begin{center}
  \begin{tabular}{|l||c|c|c|}
    \hline
    \# Tracks    &$\rho$[\%]&$\mu_t$[$\mu s$]&$\epsilon_t$[$\mu s$]\\ \hline \hline
    1 & 29.2 & 19.6 & \textbf{67.0} \\ \hline
    2 & 37.2 & 33.3 & 89.6 \\ \hline
    3 & 42.2 & 45.0 & 106.8 \\ \hline
    4 & 45.9 & 60.8 & 132.6 \\ \hline
    8 & 56.4 & 118.1 & 209.5 \\ \hline
    16 & 67.9 & 245.3 & 361.5 \\ \hline
    
  \end{tabular}
\end{center}
\caption{Number of tracks study. The strategies are evaluated on datasets \textit{Deilvery\_area} and \textit{Facade}. Scranton problem, MLP+HC, $A_{90}$. Rows correspond to different numbers of tracks conducted after the MLP is evaluated.}
\label{tab:track-numbers-study}
\end{table}

\lstset{style=mystyle}
\begin{figure}
    \centering
    \begin{lstlisting}[escapechar=!]
class Net(nn.Module):
    def __init__(self, anchors):
        super(Net, self).__init__()
        self.fc1 = nn.Linear(!\n{20}!,!\n{100}!)
        self.relu1 = nn.PReLU(!\n{100}!, !\n{0.25}!)
        self.fc2 = nn.Linear(!\n{100}!,!\n{100}!)
        self.relu2 = nn.PReLU(!\n{100}!, !\n{0.25}!)
        self.fc4 = nn.Linear(!\n{100}!,!\n{100}!)
        self.relu4 = nn.PReLU(!\n{100}!, !\n{0.25}!)
        self.fc5 = nn.Linear(!\n{100}!,!\n{100}!)
        self.relu5 = nn.PReLU(!\n{100}!, !\n{0.25}!)
        self.fc6 = nn.Linear(!\n{100}!,!\n{100}!)
        self.relu6 = nn.PReLU(!\n{100}!, !\n{0.25}!)
        self.fc7 = nn.Linear(!\n{100}!,!\n{100}!)
        self.relu7 = nn.PReLU(!\n{100}!, !\n{0.25}!)
        self.drop3 = nn.Dropout(!\n{0.5}!)
        self.fc3 = nn.Linear(!\n{100}!,anchors+1)
    def forward(self, x):
        x = self.relu1(self.fc1(x))
        x = self.relu2(self.fc2(x))
        x = self.relu4(self.fc4(x))
        x = self.relu5(self.fc5(x))
        x = self.relu6(self.fc6(x))
        x = self.relu7(self.fc7(x))
        x = self.drop3(x)
        return self.fc3(x)
\end{lstlisting}
    \caption{Code snippet describing our MLP.}
    \label{fig:MLP_code}
\end{figure}

\vspace*{1ex}
\parbox{\linewidth}{
\noindent {\bf References}\\[1ex]
{\small
\noindent \parbox{\linewidth}{\noindent \ifArXiV [70]\fi~A.L. Yuille and T. Poggio. A generalized ordering constraint for stereo correspondence. A.I.\ Memo 777, AI Lab, MIT, 1994.}}}
\end{document}